\documentclass[letterpaper, 10 pt, conference]{ieeeconf}  

\IEEEoverridecommandlockouts                              

\usepackage{graphics} 
\usepackage{epsfig} 
\usepackage{times} 
\usepackage{amsmath} 
\usepackage{amssymb}  

\usepackage{multicol}
\usepackage[normalem]{ulem}
\usepackage{xcolor,colortbl}
\usepackage{subfigure}
\usepackage{bm}
\usepackage{caption}

\title{\LARGE \bf
Structure PLP-SLAM: Efficient Sparse Mapping and Localization using Point, Line and Plane for Monocular, RGB-D and Stereo Cameras
}

\author{
Fangwen Shu
\and
Jiaxuan Wang
\and
Alain Pagani
\and
Didier Stricker
\thanks{DFKI - German Research Center for Artificial Intelligence. E-mail: { \{first\_name\}.\{last\_name\}@dfki.de}. \textbf{Acknowledgment:} This research has been partially funded by the German BMBF project MOVEON (01IS20077) and EU project CORTEX2 (grant agreement: N° 101070192).}
}

\begin{document}

\maketitle
\thispagestyle{empty}
\pagestyle{empty}

\begin{abstract}

This paper presents a visual SLAM system that uses both points and lines for robust camera localization, and simultaneously performs a piece-wise planar reconstruction (PPR) of the environment to provide a structural map in real-time. One of the biggest challenges in parallel tracking and mapping with a monocular camera is to keep the scale consistent when reconstructing the geometric primitives. This further introduces difficulties in graph optimization of the bundle adjustment (BA) step. We solve these problems by proposing several run-time optimizations on the reconstructed lines and planes. Our system is able to run with depth and stereo sensors in addition to the monocular setting. Our proposed SLAM tightly incorporates the semantic and geometric features to boost both frontend pose tracking and backend map optimization. We evaluate our system exhaustively on various datasets, and show that we outperform state-of-the-art methods in terms of trajectory precision. The code of PLP-SLAM has been made available in open-source for the research community (https://github.com/PeterFWS/Structure-PLP-SLAM).

\end{abstract}


\section{Introduction}
\label{sec:introduction}

In human-made environments, surrounding structures can often be represented by planes and line segments. Leveraging those higher-level features in visual SLAM systems has the potential to improve camera localization performance and to generate structural maps instead of unstructured point clouds. While most of the existing systems are based only on feature points and use sparse points clouds to describe the scenes and estimate the camera poses \cite{engel2017direct, engel2014lsd, forster2014svo,  klein2007parallel, mur2017orb}, these methods face various challenges in practical application, such as lacking points in low-texture environments, or poor matching performance in changing light conditions.

A line segment is a geometric primitive that has dual relation with the point, and can therefore be used in SLAM systems as efficiently as points. PL-SLAM systems \cite{gomez2019pl, pumarola2017pl} were proposed using points and lines following the pipeline of ORB-SLAM2 \cite{mur2017orb}. However, these systems adopt a representation of a 3D line with two endpoints, which can lead to ambiguities on the line direction and thus on line description (see Fig. \ref{fig: illustration_line} (a)) and suffer from occlusions or misdetections. In an attempt to solve this problem, mono PL-SLAM \cite{pumarola2017pl} followed the idea of EPnPL \cite{vakhitov2016accurate} by forcing the endpoint correspondences with a two-step procedure when minimizing the reprojection error. However,  this two-step procedure is not compatible with  optimizations based on graph optimization frameworks (e.g. g2o \cite{kummerle2011g}). In recent works such as Structure-SLAM \cite{li2020structure}, the optimization of the reprojection error on 3D lines is operated by optimizing two 3D endpoints, a representation that is over-parameterized. Since this over-parametrization cannot reliably reconstruct 3D lines, it is  a factor of deficiency in the system. To overcome these problems, in our framework, we represent 3D lines using Pl\"ucker coordinates which can be converted to an orthonormal representation with minimal parameterization \cite{bartoli2005structure}. In order to simplify the map visualization, 3D-2D correspondences search, and graph optimization, we maintain both endpoints and Pl\"ucker coordinates, and use different representations for different tasks.

\begin{figure*}[!t]
    \centering
    \includegraphics[width=0.7\linewidth]{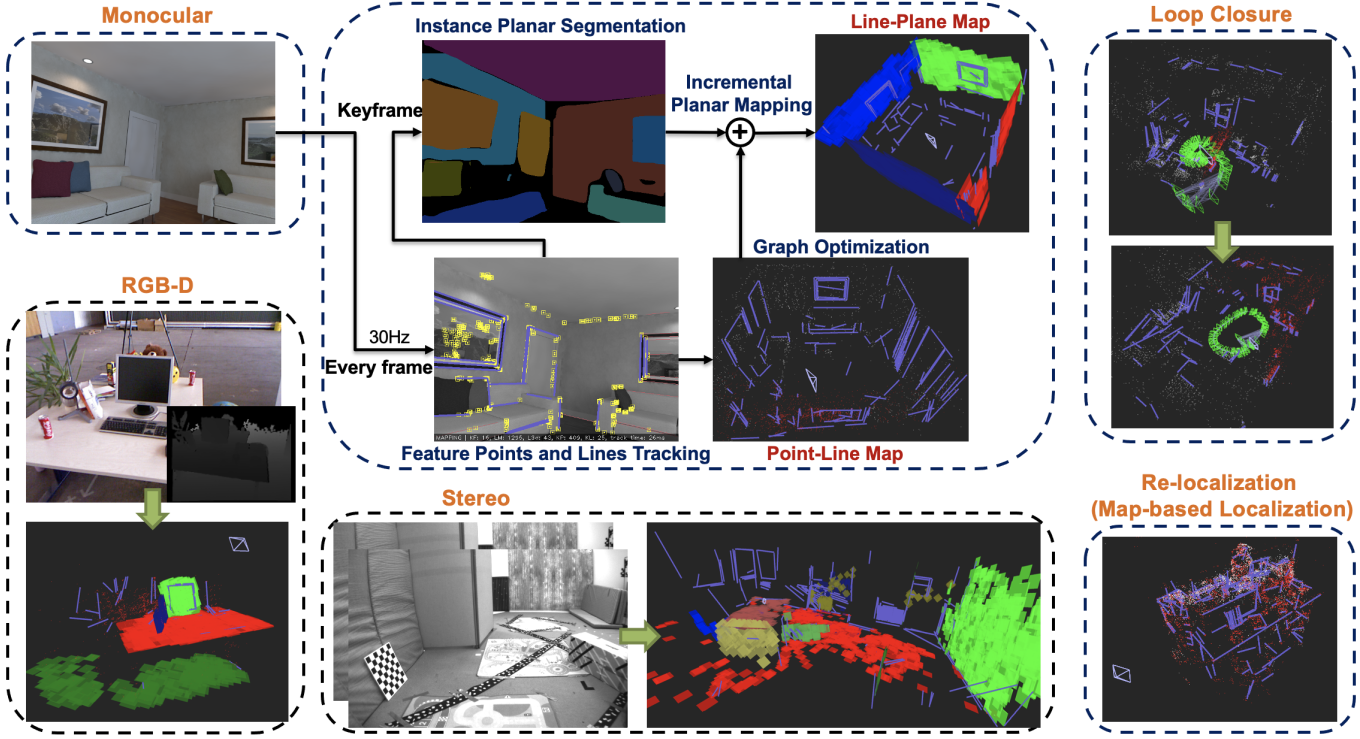}
    \caption{\textbf{The proposed SLAM system} utilizes point and line cloud for robust camera localization, with piece-wise planar reconstruction for semantic mapping. Here, the reconstructed maps are presented selectively for better visualization.}
    \label{fig: slam-workflow}
\end{figure*}

Planes, especially for indoor environments, are predominant structures that are less affected by measurement noise, as planes can be extracted e.g. from the depth image \cite{zhang2019point}. Hence, planes are often used in RGB-D SLAM \cite{hosseinzadeh2018structure, hosseinzadeh2017sparse, kaess2015simultaneous, kim2018linear, lee2017joint} with an ICP-like registration and optimization, or constraints such as Manhattan World (MW) \cite{concha2014manhattan, flint2011manhattan, yunus2021manhattanslam}. Another benefit of using planar structures is that large areas can be mapped very efficiently and used as a semantic indicator to enable intuitive and useful AR applications \cite{salas2014dense}. However, very limited works employ monocular SLAM as the backbone when considering planes due to scale ambiguity and the difficulty of fitting planes without depth sensors. In our case, following the work of Pop-up SLAM \cite{yang2016pop} which segments the ground plane using a neural network, we use an instance planar segmentation CNN \cite{liu2019planercnn, xie2021planesegnet, xie2021planerecnet} to generate a plane prior, then we incrementally reconstruct and refine the 3D piece-wise planar structure using directly the sparse point cloud generated by the monocular SLAM.

The contributions of our structure PLP-SLAM algorithm are threefold: (1) We first propose a modularized multi-feature monocular SLAM system that exploits line detection, tracking and mapping, real-time piece-wise planar reconstruction, and joint graph optimization in addition to the standard feature points. (2) We show that the loop closure can be efficiently done with a corrected line map and propose a re-localization module based on the pre-built point-line map. (3) We propose an extension of our system for RGB-D and stereo cameras. This makes our framework versatile and sensor-agnostic while tightly incorporating semantic features in SLAM. In addition, our SLAM framework is designed to be robust to noisy input in order to reduce the dependency on predictions of the CNN-based semantic planar detection. In practice,  the environment observed by the camera might be very diverse, and therefore the lines and planes are reconstructed and optimized without the need of any strong assumption such as MW. Since points, lines and planes are tightly integrated as basic features, our system is not limited to small-scale scenes. 
We benchmark our SLAM exhaustively on indoor datasets TUM RGB-D \cite{sturm12iros}, ICL-NUIM \cite{handa2014benchmark}, and EuRoC MAV \cite{Burri25012016} showing superior quantitative results compared to other state-of-the-art SLAM systems. The workflow of our SLAM framework is illustrated in Fig. \ref{fig: slam-workflow}, qualitatively showing that our reconstructed maps are accurate, intuitive, and semantically meaningful.

\section{Method}
\label{sec:implementation}

In this section, we describe our monocular SLAM system built upon OpenVSLAM \cite{sumikura2019openvslam} which is a derivative of ORB-SLAM2 \cite{mur2017orb} with high usability and extensibility. Then we highlight our contributions and the added modules.

\subsection{Structure-based Monocular SLAM}
\label{sec: integration}

One of the central modules of a SLAM system is the joint optimization of the map and the camera poses in bundle adjustment (BA). Since we are aiming at integrating higher-level features in addition to feature points, we need a proper and efficient representation of 3D lines as well as the corresponding Jacobian matrix  for iterative optimization. We integrate the planes, by adding a constraint only between 3D points and 3D planes, because the definition of a reprojection error for planar primitives is ambiguous due to their invariance to certain translations and rotations.


\subsubsection{\textbf{Exploiting Line Segments}} 
\label{sec: exploiting_line_segments}

We extract 2D line segments using LSD \cite{von2012lsd} and match them across frames via LBD descriptors \cite{zhang2013efficient}. The parameters of LSD are optimized for the best trade-off between computation efficiency and accuracy via a hidden parameter tuning and length rejection strategy, as reported in PL-VINS \cite{fu2020plvins}. Following the same idea, we achieve line segment extraction which is 3 times faster than the original implementation from OpenCV. We then use two types of representations for the line segments:

\textbf{Simple Representation of Two Ending Points.} For the visualization of a 3D line segment in the map, we use the two endpoints. This brings the advantage of a trivial projection of the 3D endpoints on the image plane, which can be used for fast retrieval of 3D-2D line matches using LBD descriptors. It also allows us to track partially occluded lines, even when one the of endpoints falls outside of the image frustum. It is important to note that the choice of the 3D endpoints has no influence on the non-linear optimization since we use another 3D line representation for the BA, as introduced in the following.

\begin{figure}[!t]
\centering
    \subfigure[Endpoints representation.]{\includegraphics[width=0.4\linewidth]{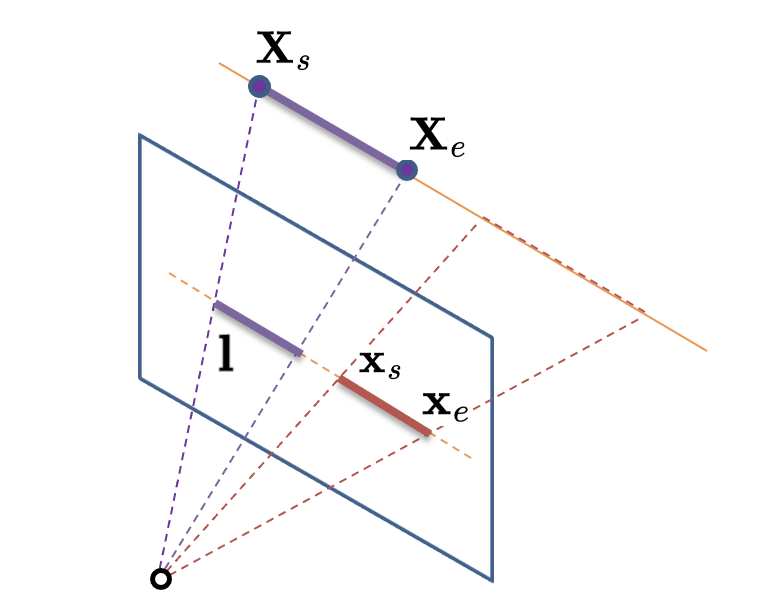}}
    \subfigure[Pl\"ucker coordinates.]{\includegraphics[width=0.4\linewidth]{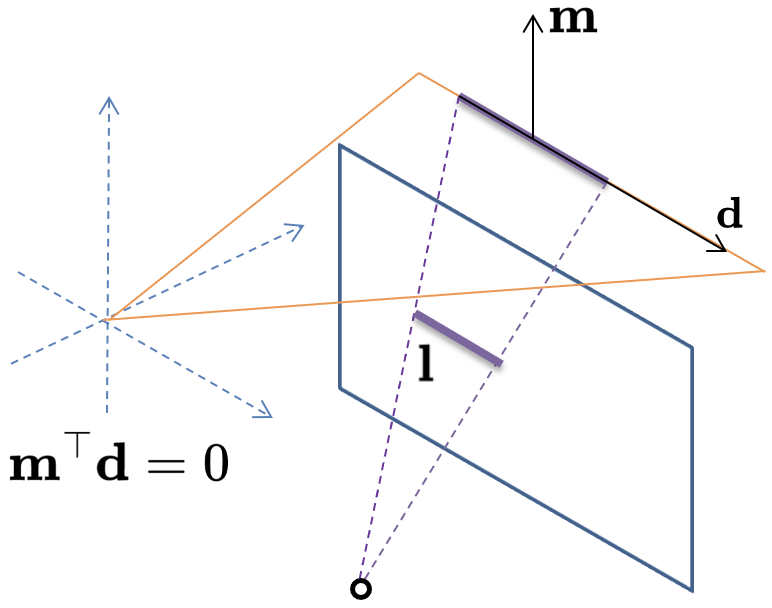}}
    \subfigure[Two-view triangulation.]{\includegraphics[width=0.37\linewidth]{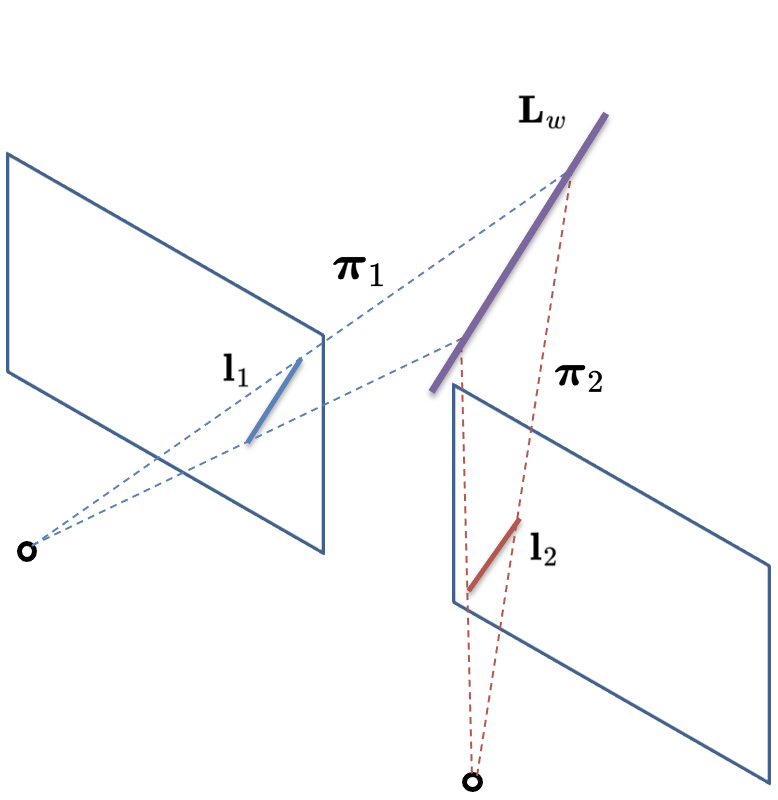}}
    \subfigure[Endpoints trimming.]{\includegraphics[width=0.37\linewidth]{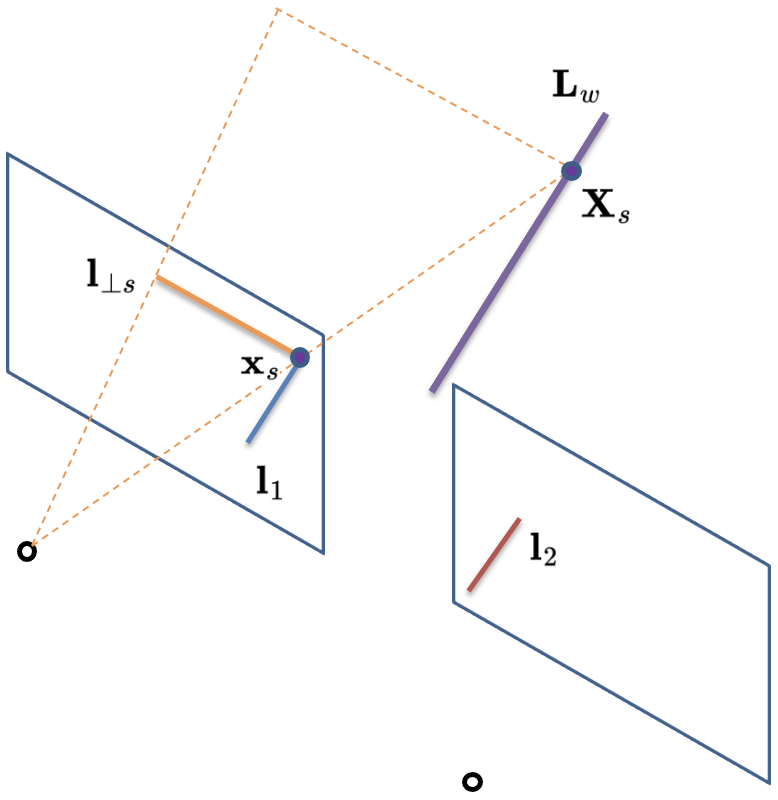}}
\caption{\textbf{The 3D line} representations and reconstruction.}
\label{fig: illustration_line}
\end{figure}

\textbf{Pl\"ucker Coordinates and Orthonormal Representation.} A 3D line can be represented with the 6D vector of Pl\"ucker coordinates as $ \mathbf{L} = (\mathbf{m}^{\top}, \mathbf{d}^{\top})^{\top}$, where the vector $\mathbf{m} \in \mathbb{R}^3$ (also called as the moment vector) is the normal to the interpretation plane containing the line $\mathbf{L}$, and $\mathbf{d} \in \mathbb{R}^3$ indicates the line direction, see Fig. \ref{fig: illustration_line} (b). Notice that it is an infinite 3D line representation, where $\mathbf{m}$ and $\mathbf{d}$ do not need to be unit vectors in the implementation.  In this way, a 3D line can be transformed from world coordinates to camera coordinates via a transformation matrix similar to 3D points, and then be projected on the image plane:

\begin{equation}
    \mathbf{L}_{c}=
    \begin{bmatrix}
    \mathbf{m}_{c} \\ \mathbf{d}_{c} 
    \end{bmatrix} 
    = \mathbf{T}_{cw} \mathbf{L}_{w} =
    \begin{bmatrix}
    \mathbf{R}_{cw} & \left[\mathbf{t}_{cw}\right]_{\times}\mathbf{R}_{cw} \\
    \mathbf{0} & \mathbf{R}_{cw} 
    \end{bmatrix}
    \begin{bmatrix}\mathbf{m}_{w} \\ \mathbf{d}_{w} 
    \end{bmatrix}
\end{equation}

\begin{equation}
\mathbf{l} = [l_{1},l_{2},l_{3}]^{\top} = \mathbf{K}_{L}\mathbf{m}_{c},
\end{equation}

\noindent where the $\mathbf{R}_{cw} \in SO(3)$ and $\mathbf{t}_{cw} \in \mathbb{R}^{3}$ are the standard rotation and translation of the camera pose. The $\mathbf{K}_{L}$ is the intrinsic matrix used to project 3D line on the image plane:

$$
\mathbf{K}_{L} =
\begin{bmatrix}
f_{y} & 0 & 0 \\
0 & f_{x} & 0 \\
-f_{y}c_{x} & -f_{x}c_{y} &  f_{x}f_{y}
\end{bmatrix}
$$

Hence, we are able to define the reprojection error of the 3D line $\mathbf{L}_{w}$ with its 2D correspondence in the image: 

\begin{equation}
\mathbf{e_{l}}= 
\left[ 
\frac{{\mathbf{x}_{s}}^{\top}\mathbf{l}}{\sqrt{l_{1}^{2}+l_{2}^{2}}} , 
\frac{{\mathbf{x}_{e}}^{\top}\mathbf{l}}{\sqrt{l_{1}^{2}+l_{2}^{2}}}
\right]^{\top}
\label{e: linereprojecterror}
\end{equation}

\noindent where $\{\mathbf{x}_{s}, \mathbf{x}_{e}\}$ are the 2D starting point and ending point of an extracted line segment from LSD.

This error term is used in BA for optimizing 3D lines and camera poses. However, Pl\"ucker coordinates are still over-parameterized, as it presents 3D line with 5-DOF in homogeneous coordinates satisfying the Klein quadric constraints $\mathbf{m}^{\top}\mathbf{d} = 0$ \cite{hartley2003multiple}. For updating the Pl\"ucker coordinates during the iterative optimization, we convert them to the minimal 4-DOF orthonormal representation \cite{bartoli2005structure}, and convert back after the optimization step.
The conversion between orthonormal representation and Pl\"ucker coordinates is given in \cite{fu2020plvins, zhang2015building}.
We implement the bundle adjustment using g2o \cite{kummerle2011g}.

\textbf{Two-view Triangulation of Lines.}
Reconstructing a line $\mathbf{L}_w$ in 3D can be achieved by forward projecting matched 2D line segments $\mathbf{l}_1$ and $\mathbf{l}_2$ from two image views to give two 3D planes, and intersecting these two planes:

\begin{equation}
    \boldsymbol{\pi}_{1} = \mathbf{l}_{1}^{\top}\mathbf{P}_{1}, \boldsymbol{\pi}_{2} = \mathbf{l}_{2}^{\top}\mathbf{P}_{2}
\end{equation}

\noindent where $\mathbf{P}_{1}$ and $\mathbf{P}_{2}$ are the standard 3$\times$4 camera projection matrices. 2D lines are constructed via their two endpoints $\{\mathbf{x}_{s}, \mathbf{x}_{e}\}$ using cross product: $\mathbf{l} = \mathbf{x}_{s} \times \mathbf{x}_{e}$. After that, the Pl\"ucker coordinates $(\mathbf{m}^{\top} , \mathbf{d}^{\top})^{\top}$ can be extracted from the dual Pl\"ucker matrix $\mathbf{L^{*}}$ \cite{hartley2003multiple}, which has the properties of:

\begin{equation}
    \mathbf{L^{*}} = \boldsymbol{\pi}_{1} \boldsymbol{\pi}_{2}^{\top} - \boldsymbol{\pi}_{2}  \boldsymbol{\pi}_{1}^{\top} =
    \begin{bmatrix}
    [\mathbf{d}]_{\times} & \mathbf{m} \\
    -\mathbf{m}^{\top} & \mathbf{0}
    \end{bmatrix}
\end{equation}

However, the triangulated 3D line is an infinite line from two intersected 3D planes, see Fig. \ref{fig: illustration_line} (c). In order to estimate the 3D endpoints for visualization and matching, using the method of endpoints trimming as discussed below.

\textbf{Endpoints Trimming and Outlier Rejection.} 
The endpoints of a 3D line can be estimated using the corresponding 2D line segment from its reference keyframe, as illustrated in Fig. \ref{fig: illustration_line} (d). 
Endpoints trimming was introduced in \cite{lee2019elaborate, zhang2015building} as a standalone method only for visualization.  In this work, we further utilize the endpoints trimming within the iterative local BA for outlier rejection and map culling, in addition to the $\chi^{2}$ distribution test on the reprojection error. It is integrated as part of the positive depth checking (such as $Z_{c} > 0$), and evaluates if the absolute change of the position $\Delta X$ of the 3D endpoints compared to the median depth of the scene is less than a ratio (0.1) after optimization finished. If it is not the case, the 3D line is an outlier (possibly from mismatching or triangulated with two-view ambiguity when the 3D line is located close to the epipolar plane). Experimentally, we found that this makes the BA more efficient and robust to outliers, the reconstructed line cloud remains accurate, compact, and clean, which can be observed from Fig. \ref{fig: slam-workflow} and Fig. \ref{fig: qualitative-map}. Moreover, endpoints trimming is also used to correct the line map during the loop closure (see Sec. \ref{sec: loop_detection_closure_global_BA}).


\subsubsection{\textbf{Exploiting 3D Planar Structures}} 
\label{sec: exploiting_planar_tructures}

\begin{figure}[!t]
\centering
    \subfigure[3D plane reconstruction with spatial coherence.]{\includegraphics[width=0.43\linewidth]{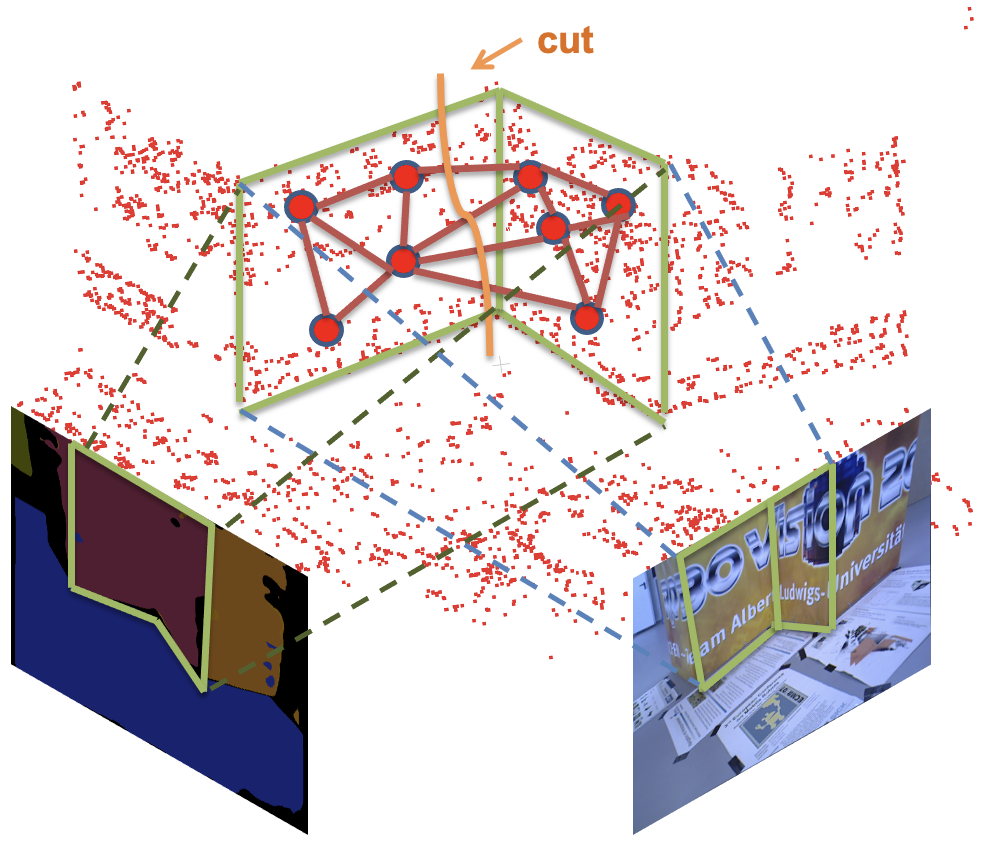}}
    \hfill
    \subfigure[The possible distribution of 3D map points around a plane \cite{lee2011mav}.]{\includegraphics[width=0.46\linewidth]{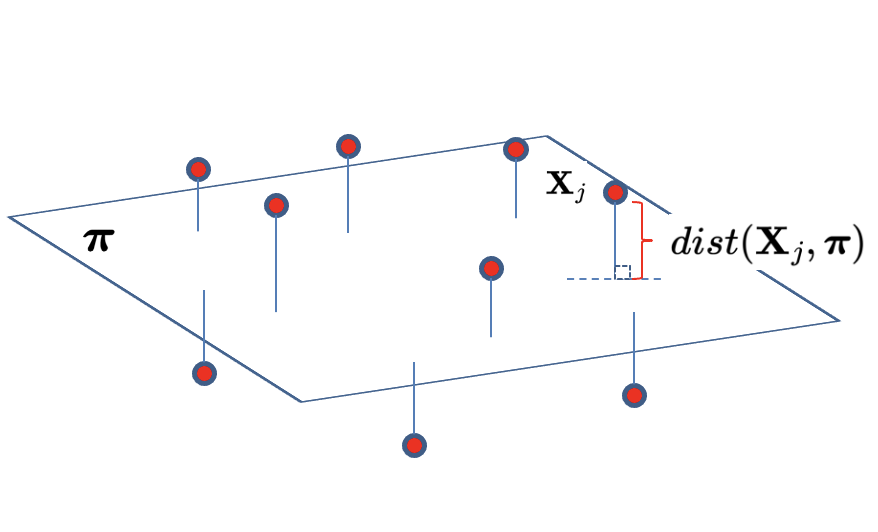}}\caption{\textbf{The interpretation of 3D plane:} (a) An example of 3D plane fitting optimized with spatial coherence. We initialize the plane by the given instance planar segmentation. Note that if two planes are wrongly segmented as one (e.g. due to texture similarity), they are separated later thanks to our graph-cut within a local neighborhood graph in 3D, as highlighted. (b) The possible distribution of the 3D points.}
\label{fig: illustration_plane}
\end{figure}

The reconstruction of 3D planes in a monocular SLAM is challenging because of the limited 3D information. At the same time, the Manhattan World (MW) assumption does not necessarily hold if we are trying to reconstruct all the possible plane instances observed from a single image (not limited to the perpendicular layout).
In order to reconstruct planes, we leverage the pairwise relationship between the 3D points and 3D planes in SLAM backend, and utilize the geometric relationship to minimize the 3D distance, as illustrated in Fig. \ref{fig: illustration_plane} (b). Here, we adopt the infinite plane representation $\boldsymbol{\pi} = (\mathbf{n}^{\top}, d)^{\top}$ \cite{kaess2015simultaneous}, where $\mathbf{n}$ is the plane normal and $d$ is the distance to the world origin. 

A 3D plane instance is reconstructed by fitting a set of sparse and noisy 3D points triangulated in real-time. We initialize the set of points belonging to a single plane by predicting the instance planar segmentation of the input images using PlaneRecNet \cite{xie2021planerecnet} (only on keyframes). However, in order to cope with possible misclassification from the neural network, especially on the unseen data of SLAM benchmarks, the reconstruction is conducted as a sequential RANSAC \cite{shu2021visual} coupled with an inner local optimization of Graph-cut \cite{barath2021graph}. In this way, we locally optimize the spatial coherent planes in 3D space, as shown in Fig. \ref{fig: illustration_plane} (a). To this aim, we formulate the plane fitting problem as an optimal binary labeling problem \cite{isack2012energy} with the energy term:

\begin{equation}
    E(\boldsymbol{\Pi}) = \sum_{v}^{} \|\Pi_v\| + \lambda \cdot \sum_{(u,v) \in \mathcal{N}}{\delta (\Pi_u \neq \Pi_v)}
    \label{e: energy_plane_fitting}
\end{equation}

\noindent where the first term $\sum_{v}^{} \|\Pi_v\|$ counts inliers for the target plane model using 0-1 measurement:

\begin{equation}
    \|\Pi_v\|_{\{0;1\}} = 
    \begin{cases}
		0 & \text{if } (\Pi_v = 1 \wedge \mathrm{dist}(\mathbf{v}, \boldsymbol{\pi}) < \epsilon_{d}) \text{ } \vee \\    
		 & \text{\phantom{xx}} (\Pi_v = 0 \wedge \mathrm{dist}(\mathbf{v}, \boldsymbol{\pi}) \geq \epsilon_{d}) \\    
        1 & \text{otherwise.}
	\end{cases}
	\label{e: unary_energy}
\end{equation}

In Eq. (\ref{e: energy_plane_fitting}), $\boldsymbol{\Pi}=\{ \Pi | \mathbf{v} \in V \}$ is the assignment of plane models to 3D point $\mathbf{v}$, and $V$ indicates the set of 3D vertices from a neighborhood graph. Here, we use the distance between a 3D point and the plane as the geometric error measure in Eq. (\ref{e: unary_energy}): $\mathrm{dist}(\mathbf{v}, \boldsymbol{\pi})=| \frac{\mathbf{n}^{\top}\mathbf{v} + d}{\|\mathbf{n}\|} |$.

Moreover, in Eq. (\ref{e: unary_energy}) the parameter $\Pi_{v} \in \{0,1\}$ indicates the labeling. Here, the unary energy penalizes nothing when a 3D point is labeled as an inlier (close to the plane) or it is labeled as an outlier (far from the plane). The second term of Eq. (\ref{e: energy_plane_fitting}) indicates the spatial regularization \cite{isack2012energy} which penalizes neighbors with different labels in the graph. $\delta (\cdot)$ is 1 if the specified condition inside the parenthesis holds, and 0 otherwise. The neighborhood graph $ \mathcal{N}$ is constructed using the Fast Approximate Nearest Neighbors algorithm \cite{muja2009fast} according to a predefined sphere radius $r$ ($= 2\epsilon_{d}$), and the minimum samples (3 points formulate a plane) are sampled uniformly. $\lambda$ is a parameter balancing the two terms, which is set as 0.6 in our experiments. 

\textbf{Incremental Merging and Refinement.} 
In order to cope with possibly large planes, we add a merging mechanism in the local mapping thread. Two planes are merged if the following two conditions are met: first they should have nearly parallel normals: $|cos(\theta)| = |\frac{\mathbf{n_i}\mathbf{n_j}}{\|\mathbf{n_i}\|\|\mathbf{n_j}\|}| > T_{\theta}$ (set as 0.8 in this work)
and second, they should be geometrically close to each other: $|\frac{d_i}{\|d_i\|} - \frac{d_j}{\|d_j\|}| < T_{d}$.
The new plane equation is then updated according to a model residual threshold $\epsilon_{\Pi}$ in a RANSAC loop. Following this, all associated point landmarks are projected on the plane by minimizing the point-plane distance via: 
$\hat{\mathbf{v}} = \mathbf{v} - \mathrm{dist}(\mathbf{v}, \boldsymbol{\pi})\frac{\mathbf{n}}{\|\mathbf{n}\|}$.

\textbf{Adaptive Geometric Thresholds.}
Some of the parameters introduced in the previous equations need to be adjusted according to different environments. Examples of such parameters are $\epsilon_{d}$ in Eq. (\ref{e: unary_energy}), the parameter $T_{d}$ used to merge planes, and the model residual $\epsilon_{\Pi}$ used to stop the RANSAC loop. In order to avoid case-dependent parameter-tuning, we follow the work of \cite{shu2021visual} with an adaptive parameter setting strategy, where the above-mentioned thresholds will be adjusted dynamically according to the scene depth of the local map observed by the reference keyframe.


\subsection{Motion-only BA and Local BA}
\label{sec: motion-local-BA}

In this work, the reprojection errors of the points and lines are minimized in two different bundle adjustments: motion-only BA and local BA. The overall cost function is:

\begin{equation}
C =
    \sum_{i,j} \rho_{h}(\mathbf{e}_{ij}^{\top}\bm{\Omega}_{ij}\mathbf{e}_{ij}) 
    + \sum_{i,z} \rho_{h}(\mathbf{e_{l}}_{iz}^{\top}\bm{\Omega}_{iz}\mathbf{e_{l}}_{iz})
    \label{e: cost}
\end{equation}

\noindent implicitly with the minimized distance between 3D point and 3D plane, because the planes are statistically fitted in the optimal position via SVD, the associated 3D points are projected on the plane as discussed in the last Sec. \ref{sec: exploiting_planar_tructures}:

$$
    \sum_{j,k} dist(\mathbf{X}_j, \bm{\pi}_k) 
$$

\noindent where $i,j,k, z$ are the number of camera views, 3D points, 3D planes, and 3D lines, respectively. In Eq. (\ref{e: cost}), the first term indicates the standard reprojection error for feature points, and the second term indicates the line reprojection error explained by Eq. (\ref{e: linereprojecterror}). Moreover, the 6-DOF camera pose is represented as Lie algebra $\mathfrak{se}(3)$, and the 4-DOF line is represented as orthonormal representation. $\rho_{h}$ is the Huber robust cost function and $\bm{\Omega}_{ij}, \bm{\Omega}_{iz}$ are the covariance matrices associated with the scale (of image pyramid) at which feature point or line segment was detected. In this work, we only utilize the line segments extracted from the original image resolution, thus $\bm{\Omega}_{iz} = \mathbf{I}_{2\times2}$.

The analytical Jacobians for point are well-known, while the Jacobians for line can be analytically calculated by chain rule to make derivations which are given in \cite{ bartoli2005structure, lee2019elaborate, zuo2017robust}.


\begin{figure*}[!t]
    \centering
    \subfigure[(\textbf{Monocular}) Map of \textit{fr3\_structure\_texture\_far}. This example shows that our line and plane reconstruction are accurate when using monocular SLAM.]{\includegraphics[width=0.23\linewidth]{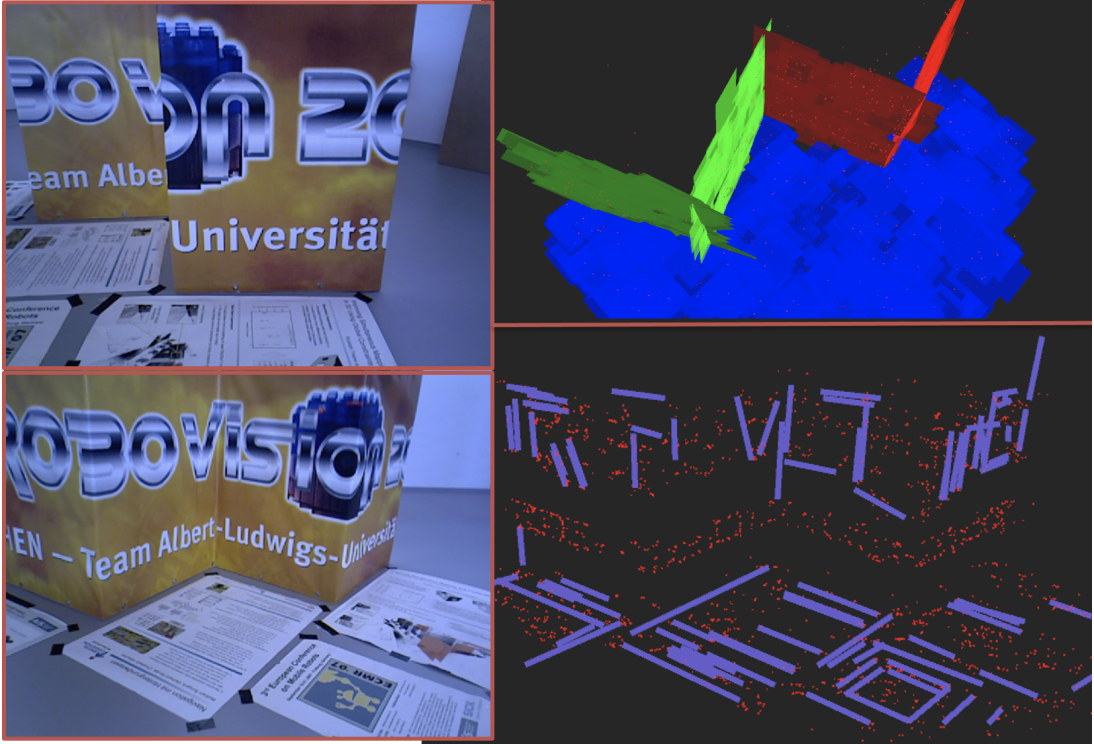}}
    \hfill
    \subfigure[(\textbf{RGB-D}) Map of \textit{office\_room\_traj0}. Here, the ceiling (red) is perfectly reconstructed, and the furniture can be observed combined with the plane (blue/green) and lines. Same for the paintwork (yellow) on the wall. This example verifies that our sparse semantic mapping is efficient and intuitive.]{\includegraphics[width=0.335\linewidth]{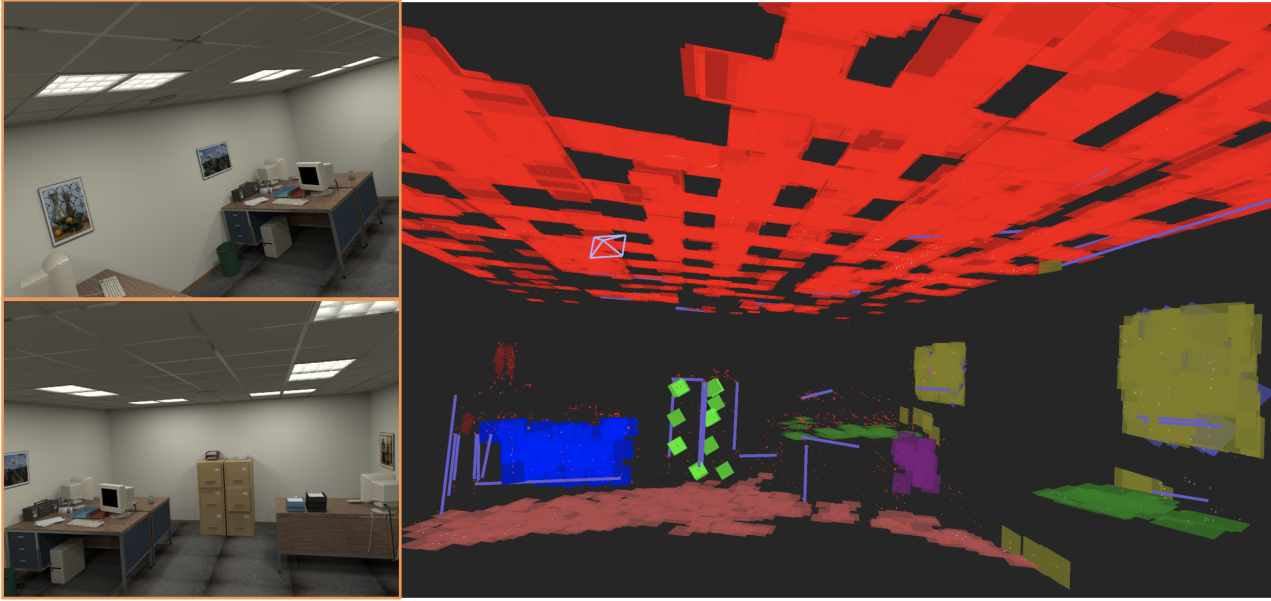}}
    \hfill
    \subfigure[(\textbf{Stereo}) Point-line map of \textit{MH\_04\_difficult}. No planar structure due to the failure of CNN on this factory sequence. However, our map is as good as the map reconstructed from PL-VINS \cite{fu2020plvins} which is a visual-inertial SLAM using points and lines. 
    ]{\includegraphics[width=0.34\linewidth]{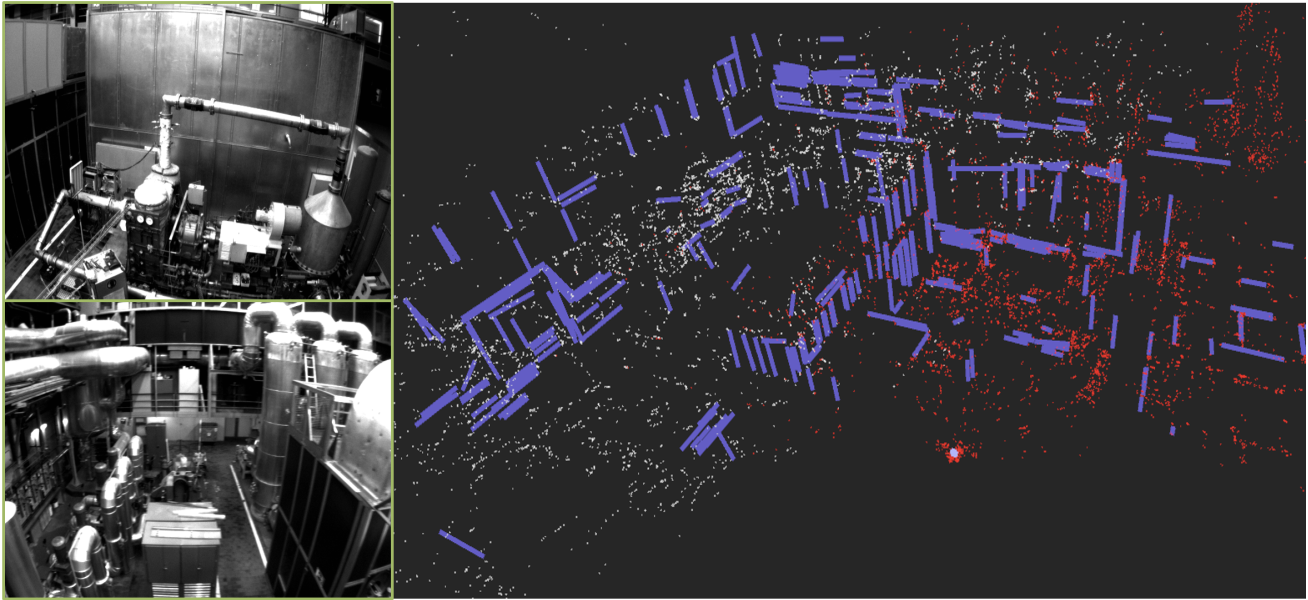}}
    \caption{\textbf{The qualitative results of various reconstructed maps} under different sensor settings, complementary to Fig. \ref{fig: slam-workflow}. 
    }
    \label{fig: qualitative-map}
\end{figure*}


\subsection{Loop Detection, Loop Closure and Global BA}
\label{sec: loop_detection_closure_global_BA}

Loop detection of monocular SLAM aims at estimating the 7-DOF similarity transformation $\mathrm{Sim(3)}$ after a best-validated loop candidate (keyframe) is found:

\begin{equation}
    \mathrm{Sim(3)} =
    \left\{ 
    \mathbf{S}_{\mathrm{point}} = 
    \begin{bmatrix}
        s\mathbf{R} & \mathbf{t} \\
        \mathbf{0} & \mathbf{1}
    \end{bmatrix} 
    \in \mathbb{R}^{4\times4}
    \right\}
\label{e: sim3}
\end{equation}

In this work, we do not address the place recognition problem by re-building a BoW (Bag of Words) vocabulary using the LBD descriptors as done in \cite{gomez2019pl}.
Instead, we use the given DBoW vocabulary \cite{galvez2012bags} built from ORB features \cite{mur2014fast} for loop detection. The 3D line similarity transformation \cite{bartoli20013d} is then calculated according to Eq. (\ref{e: sim3}):
 
\begin{equation}
    \mathbf{S}_{\mathrm{line}} = 
    \begin{bmatrix}
        s\mathbf{R} & \left[\mathbf{t}\right]_{\times}\mathbf{R} \\
    \mathbf{0} & \mathbf{R} 
    \end{bmatrix}
    \in \mathbb{R}^{6\times6}
\end{equation}

\noindent where both similarity transformations are expressed by the same scale factor, rotation and translation matrices.

Thus, we are able to correct the 3D line map (represented by the Pl\"ucker coordinates) in a similar way as the correction of a 3D point cloud within the step of Loop Fusion described in \cite{mur2015orb}. Thereafter, we optimize the Essential graph over the similarity transformations, which distributes the loop closing error along the graph and correct scale drifts \cite{strasdat2010scale}. Then the 3D map points and lines are transformed according to the correction of the reference keyframe that observed them. This procedure is illustrated in Fig. \ref{fig: slam-workflow}. To achieve the optimal solution, we perform a global BA in a separate thread \cite{mur2017orb}, and endpoints trimming is used to re-estimate 3D line endpoints and function as a map culling method.


\begin{table}[!t]
\centering
\scalebox{0.6}{
\begin{tabular}{|l||c|c|c|c|c|}
\hline
    \textbf{Monocular} & \textbf{Ours} & \textbf{Ours} & \textbf{Ours} & Structure & Object-  \\
    & (full) & & & SLAM \cite{li2020structure} & Plane \cite{yang2019monocular} \\
\hline
    Config. & point   & point   & point   & point    & point    \\
            & + line  & + line  &         & + line   &          \\
            & + plane &         & + plane & + plane  & + plane  \\
            &         &         &         & + normal & + object \\
\hline 
\hline
    living\_rm\_0  &\textbf{0.26}	&0.32	&0.39	&-	&0.80\\
    living\_rm\_2  &2.21	&\textbf{1.88}	&2.67	&4.50	&2.06\\
    living\_rm\_3  &1.95	&\textbf{1.81}	&2.54	&4.60	&5.38\\
    office\_0  &5.14	&\textbf{4.50}	&5.26	&-	&5.93\\
    office\_2  &4.07	&3.55	&5.31	&3.10	&\textbf{2.63}\\
    office\_3  &3.67	&\textbf{2.95}	&5.65	&6.50	&-\\
\hline
    \textbf{Avg.} &\underline{2.88}	&\textbf{2.50}	&3.64	&4.68	&3.36\\
\hline
\end{tabular}
}
\caption{\textbf{Monocular SLAM} evaluated on \textbf{ICL-NUIM \cite{handa2014benchmark}}, presented are the \textbf{absolute trajectory errors (ATEs) RMSE [cm]} (- stands result not available). Each result from ours was calculated as the average over 5 executions. 
}
\label{tab: ATE-icl-mono}
\end{table}


\subsection{Re-localization}
\label{sec: re-localization}

The existing method in feature-based SLAM utilizes the global descriptor of BoW \cite{galvez2012bags} for image retrieval, thereafter using the $O(n)$ closed-form solution of EPnP \cite{lepetit2009epnp} to initialize the iterative optimization, as the run-time requirement is critical. Therefore, simply replacing the EPnP with EPnPL \cite{vakhitov2016accurate} as done in mono PL-SLAM \cite{pumarola2017pl} brings no significant improvement. In this work, we use a BA with both point and line reprojection errors that provide better-refined camera poses. Note that we optimize over orthonormal representation of the line, instead of forcing the endpoints correspondence as done in \cite{pumarola2017pl} (in the spirit of EPnPL), which means that our method is naturally more efficient and avoids the shifting ambiguity of the line during nonlinear optimization.  

\section{Experiments and Results}
\label{sec:results}

We report our experiments on the datasets TUM RGB-D \cite{sturm12iros} and ICL-NUIM \cite{handa2014benchmark} for monocular and RGB-D SLAM.
We only present the qualitative evaluation on EuRoC MAV \cite{Burri25012016} due to limited pages, please refer to Fig. \ref{fig: slam-workflow} and Fig. \ref{fig: qualitative-map}.


\begin{table}[!t]
\centering
\scalebox{0.6}{
\begin{tabular}{|l||c|c|c|c|c|c|}
\hline
    \textbf{Monocular} & \textbf{Ours} & \textbf{Ours} & \textbf{Ours}  & PL-SLAM  & Elaborate & Structure \\
    & (full) & & & \cite{pumarola2017pl} &  \cite{lee2019elaborate} & SLAM \cite{li2020structure}\\
\hline
    Config. & point   & point   & point     & point  & point   & point  \\
                  & + line  & + line  &           & + line & + line  & + line \\
                  & + plane &         & + plane   &        &         & + plane \\
                  &         &         &           &        &         & + normal \\
\hline
\hline
    fr1\_xyz    & 1.06 & 1.05 & 1.09 & 1.21 & \textbf{1.02} & -\\
    fr1\_floor  & 2.03 & 2.24 & \textbf{1.85} & 7.59 & 3.49 & -\\ 
    fr1\_desk   & 2.02 & \textbf{1.65} & 1.82 & - & - & -\\
    fr2\_xyz    & 0.26 & 0.26 & \textbf{0.25} & 0.43 & 0.31 & -\\
    fr2\_desk   & 0.94 & \textbf{0.77} & 1.14 & - & 4.65 & -\\
    fr3\_st\_tex\_far     & 0.99 & 1.11 & 1.05 & 0.89 & \textbf{0.87} & 1.40\\
    fr3\_st\_tex\_near    & \textbf{1.14} & 1.44 & 1.15 & 1.25 & - & 1.40\\
    fr3\_nst\_tex\_near   & \textbf{1.26} & 1.41 & 1.48 & 2.06 & - & -\\
    fr3\_nst\_tex\_far    & \textbf{3.24} & 3.37 & 3.51 & - & 3.68 & -\\
    fr3\_long\_office     & 1.16 & \textbf{1.04} & 1.54 & 1.97 & 2.98 & -\\
\hline
    \textbf{Avg.} & \textbf{1.41} & \underline{1.43} & 1.49 & 2.20 & 2.43 & - \\
\hline
\end{tabular}
}
\caption{\textbf{Monocular SLAM} evaluated on dataset \textbf{TUM RGB-D} \cite{sturm2012benchmark}, presented are the \textbf{ATEs RMSE [cm]}.
}
\label{tab: ATE-tum-mono}
\end{table}

\subsection{Performance of Visual SLAM System}
\label{sec: slam_performance}

\textbf{Effectiveness of line segments.} We present the trajectory errors of our monocular SLAM compared to other state-of-the-art systems. The evaluated dataset ICL-NUIM provides low-contrast and low-texture synthetic indoor image sequences which are very challenging for monocular SLAM, while the TUM RGB-D dataset provides real-world indoor sequences under different texture and structure conditions. The effectiveness of adding line segments with a well-formulated reprojection error in SLAM can be found in the 3rd column (\textbf{ours}: point + line) of Table \ref{tab: ATE-icl-mono}. We observe that line segments improve the performance of monocular SLAM remarkably in the case of lacking feature points under low-texture and low-contrast environments. It also brings SLAM a more robust camera localization performance when there are enough point features tracked (on TUM RGB-D dataset of Table \ref{tab: ATE-tum-mono}), and similar results can be observed from Table \ref{tab: ATE-icl-rgbd} and Table \ref{tab: ATE-tum-rgbd} with RGB-D setting.

\textbf{The implicit constraint from 3D planar structures.} In Table \ref{tab: ATE-tum-mono}, we observe that our implicit constraint that minimizes the distance between the 3D points and planes brings performance improvement to monocular SLAM on pure planar scenes such as \textit{fr1\_floor}, \textit{fr3\_structure\_texture\_near} and \textit{fr3\_nostructure\_texture\_far}, \textit{fr3\_nostructure\_texture\_near\_with\_loop}.
Moreover, in Table \ref{tab: ATE-icl-rgbd} where we evaluate our RGB-D SLAM, when the depth sensor is available, this simple and efficient constraint also regularizes the point cloud (associated with certain planes) triangulated from the depth image, which results in the best average performance (\textbf{ours}: point + line + plane). This point-plane distance constraint is theoretically possible to be integrated into graph optimization \cite{lee2011mav} with a unary edge linked to the 3D point vertex, while the position of the plane is considered statistically optimal from RANSAC and SVD, hence fixed during optimization. Practically, this is equivalent to what we implemented via a simple projection of the 3D point on the plane. We experimentally found out that adding this constraint into the pose-graph will disturb the optimization procedure, because the local BA then becomes a large-scale non-linear optimization problem since such unary edges do not directly constrain the camera poses, and certain planes (e.g. floor) may dominate the map and violates the local optimization strategy. 

\textbf{Failure cases.} We observed that, e.g., monocular SLAM is not able to initialize on sequence \textit{living\_room\_1} (of ICL-NUIM) which can be solved by adjusting the initialization threshold. On sequence \textit{office\_room\_1} the monocular SLAM will lose its tracking due to brutal movement when the camera targets a no-texture corner of the room. This happens to all other monocular SLAM systems, so we removed these two sequences from evaluation for consistency (in Table \ref{tab: ATE-icl-mono}). Another factor of failure in our framework is semantic planar mapping, which will be discussed in the next section.

\begin{figure}[!t]
\centering
    \includegraphics[width=0.9\linewidth]{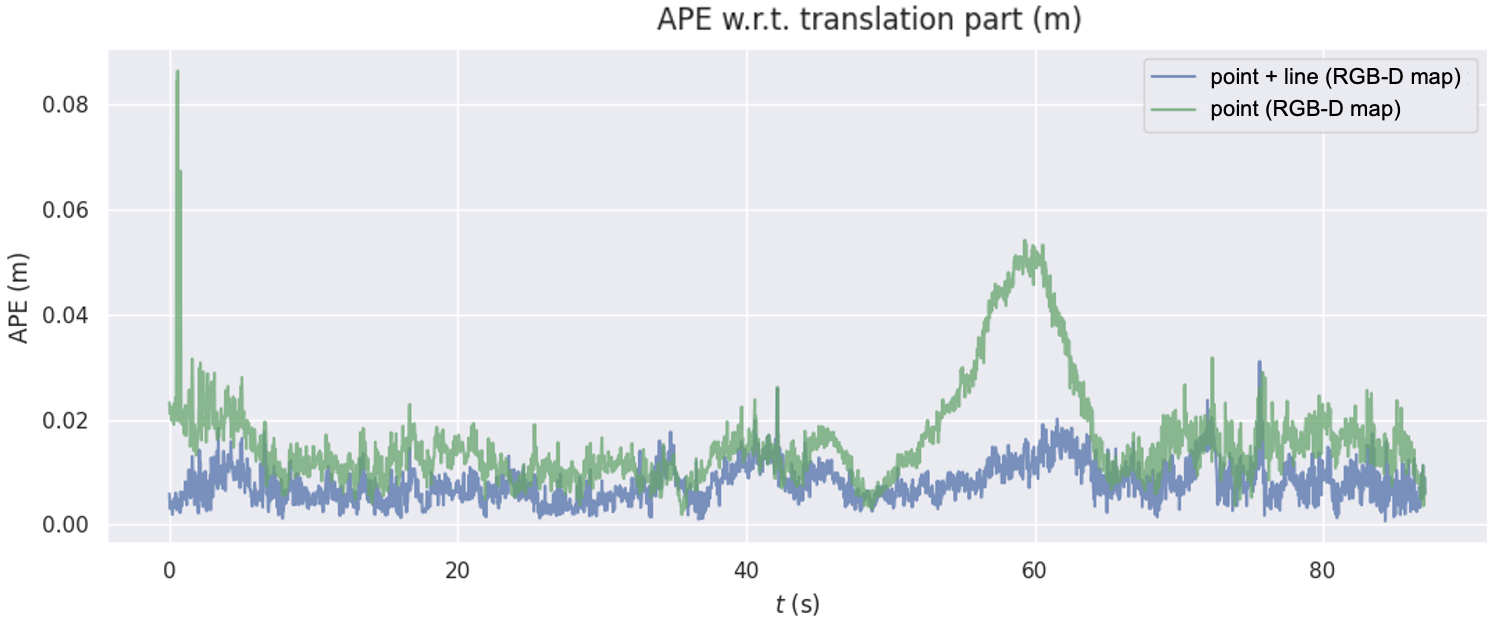}

\caption{\textbf{The evaluation of re-localization module.} We illustrate the absolute pose error (APE) [m] of all the relocalized images from sequence \textit{freiburg3\_long\_office\_household}, where the x-axis indicates the relative timestamps.
}
\label{fig: ape_reloc}
\end{figure}

\subsection{Sparse Semantic Mapping}
In Fig. \ref{fig: slam-workflow} and \ref{fig: qualitative-map}, we illustrate reconstructed point-line and line-plane maps respectively. Our map representation is designed as a lightweight sparse map, which intuitively shows the scene structure. The 3D planar structure is visualized using a rectangular plane-patch centered around the associated 3D point. In this way, we make use of the non-structure point cloud for visualizing the structural plane, efficiently without added computation. 3D lines are naturally observable, which illustrates most edge features of the scene. 

3D plane reconstruction may fail, when the CNN-based segmentation is too noisy or even fails on ambiguous images. Thus, we use a graph-based optimization in RANSAC to fit planes. However, it is not avoidable that some user-defined thresholds of RANSAC are difficult to fine-tune. We solve this partially by introducing our adaptive parameter setting strategy \cite{shu2021visual}. One drawback of our planar map representation is that it strongly depends on the map points, which brings the limitation that no reliable plane can be fitted when there are not enough point landmarks. Our map refiner only keeps high-quality planes, which also omits smaller planes from the map. This results in mapping failure for low-texture scenes. Notice that the failure of planar mapping will not negatively influence the tracking thread of SLAM using points and lines, however, we remove the results from ours in Table \ref{tab: ATE-tum-rgbd} if the plane reconstruction failed.

\subsection{Re-localization Module}
\label{sec: ape_reloc}

Our re-localization module estimates the camera pose without using any prior information other than the map. This is useful when the previous camera location cannot be used, such as when the tracking is lost. This problem was recently addressed as the map-based visual localization problem \cite{gao2021pose, yu2020line, yu2020monocular} or global localization \cite{peng2021megloc, sarlin2019coarse}. We evaluate our re-localization module via conducting the single monocular image map-based localization, which is based on a pre-built map from our RGB-D SLAM system (as shown in Fig. \ref{fig: slam-workflow} bottom-right, of sequence \textit{freiburg3\_long\_office\_household}). Thus, given every single image from this data sequence, we estimate the initial camera pose via image retrieval followed by EPnP, and then optimize it using motion-only BA with 3D-2D point and line correspondences, as discussed in Sec. \ref{sec: re-localization}.  To this end, we calculate the absolute camera position errors (APEs) using all relocalized images. As shown in Fig. \ref{fig: ape_reloc}, where the blue line indicates using points and lines for pose optimization, which shows clearly smaller errors compared to the one (green line) using points only.


\begin{table}[!t]
\centering
\scalebox{0.52}{
\begin{tabular}{|l||c|c|c|c|c|c|}
\hline
    \textbf{RGB-D} & \textbf{Ours} & \textbf{Ours} & \textbf{Ours} & Manhattan & Structural  & SP-SLAM  \\
    & (full) & & & SLAM \cite{yunus2021manhattanslam} & RGB-D \cite{li2021rgb} & \cite{zhang2019point} \\
\hline
    Config. & point   & point   & point   & point    & point    & point  \\
            & + line  & + line  &         & + line   & + line   &         \\
            & + plane &         & + plane & + plane  & + plane  & + plane  \\
\hline 
\hline
    living\_rm\_0  &\textbf{0.54}	&0.63	&0.58	&0.70	&0.60	&0.80\\
    living\_rm\_2  &\textbf{1.26}	&1.46	&1.36	&1.50	&2.00	&1.92\\
    living\_rm\_3  &\textbf{0.68}	&1.07	&1.16	&1.10	&1.20	&1.25\\
    office\_0  &\textbf{1.91}	&1.99	&2.31	&2.50	&4.10	&1.99\\
    office\_1  &2.66	&2.26	&2.20	&\textbf{1.30}	&2.00	&2.25\\
    office\_2  &0.83	&\textbf{0.79}	&0.89	&1.50	&1.10	&2.20\\
    office\_3  &\textbf{0.84}	&0.96	&0.86	&1.30	&1.40	&1.84\\
\hline
    \textbf{Avg.} &\textbf{1.25}	&\underline{1.31}	&1.34	&1.41	&1.77	&1.75\\
\hline
\end{tabular}
}
\caption{\textbf{RGB-D SLAM} evaluated on dataset \textbf{ICL-NUIM \cite{handa2014benchmark}}, presented are the \textbf{ATEs RMSE [cm]}.
}
\label{tab: ATE-icl-rgbd}
\end{table}
\begin{table}[!t]
\centering
\scalebox{0.6}{
\begin{tabular}{|l||c|c|c|c|c|}
\hline
    \textbf{RGB-D} & \textbf{Ours} & \textbf{Ours} & \textbf{Ours}  & Manhattan & SP-SLAM \\
    & (full) & &  & SLAM\cite{yunus2021manhattanslam} & \cite{zhang2019point} \\
\hline
    Config. & point   & point   & point     & point   & point   \\
            & + line  & + line  &           & + line  &         \\
            & + plane &         & + plane   & + plane & + plane \\
\hline
\hline
    fr1\_xyz    &1.03	&0.98	&1.10	&1.00	&\textbf{0.93}\\
    fr1\_floor  &\textbf{1.21}	&1.40	&1.36	&-	&-\\
    fr1\_desk   &2.02	&1.89	&1.87	&2.70	&\textbf{1.43}\\
    fr2\_xyz    &1.45	&1.68	&1.42	&\textbf{0.80}	&-\\
    fr3\_st\_tex\_far     &\textbf{0.89}	&0.99	&0.98	&2.20	&0.97\\
    fr3\_st\_tex\_near    &1.00	&1.05	&0.95	&1.20	&\textbf{0.84}\\
    fr3\_nst\_tex\_near   &1.59	&\textbf{1.19}	&1.64	&-	&-\\
    fr3\_nst\_tex\_far    &3.52	&\textbf{3.05}	&4.35	&-	&-\\
    fr3\_long\_office     &1.01	&\textbf{0.91}	&0.94	&-	&-\\
    fr3\_large\_cabinet &4.99	&4.37	&5.03	&8.30	&\textbf{2.97}\\
    fr3\_st\_notex\_far &-	&\textbf{1.46}	&-	&4.00	&2.20\\
    fr3\_st\_notex\_near &-	&\textbf{0.98}	&-	&2.30	&1.25\\
\hline
    \textbf{Avg.} &1.87	&\underline{1.66}	&1.96	&2.81	&\textbf{1.51}\\
\hline
\end{tabular}
}
\caption{\textbf{RGB-D SLAM} evaluated on dataset \textbf{TUM RGB-D} \cite{sturm2012benchmark}, presented are the \textbf{ATEs RMSE [cm]}. 
}
\label{tab: ATE-tum-rgbd}
\end{table}

\section{Conclusion}
\label{sec:conclusion}

In this work, we presented a sparse visual SLAM method that uses both points and line segments for robust camera localization, together with a plane detection and reconstruction mechanism which increases both pose robustness and map semantic interpretation.
Our comprehensive evaluations show high qualitative results of the reconstructed maps, and better quantitative performance in terms of pose estimation compared to state-of-the-art SLAM algorithms. As the system was designed based on monocular assumption, future work would be generalizing this full SLAM system to more challenging scenarios, such as scenes with low textures. We also plan to extend the system to other high-level features such as 3D objects in the scene.











\bibliographystyle{ieee}
\bibliography{egbib}

\clearpage
\section{Supplementary Materials}


\subsection{Hidden Parameters of LSD Extractor}
\label{app: hidden_parameters_LSD}

\begin{table}[!h]
    \centering
\scalebox{1.0}{
    \begin{tabular}{l|c}
    Name of the Parameter & Value \\
    \hline
    opts.refine & 1  \\
    opts.scale & 0.5  \\
    opts.sigma\_scale & 0.6  \\
    opts.quant & 2.0  \\
    opts.ang\_th & 22.5  \\
    opts.log\_eps & 1.0  \\
    opts.density\_th & 0.6  \\
    opts.n\_bins & 1024  \\
    opts.min\_length & 0.125 * min(img.cols, img.rows)  \\
    \hline
    \end{tabular}
}
    \caption{\textbf{The hidden parameters used for extracting 2D line segments using LSD}, as discussed in PL-VINS \cite{fu2020plvins}, which can be 3 times faster than the original implementation in OpenCV. For the meaning of the parameters please refer to the OpenCV documentation of LSD.}
    \label{tab: lsd_parameters}
\end{table}




\subsection{Conversion between Orthonormal Representation and Pl\"ucker Coordinates}
\label{app: Conversion}

As an infinite line representation, the Pl\"ucker coordinates $\mathbf{L}_{w}=(\mathbf{m}_{w}^{\top},\mathbf{d}_{w}^{\top})^{\top}$ can be calculated by intersecting two 3D planes as discussed in two-view triangulation or constructed by two 3D endpoints $\mathbf{X_{1}} = (X_{1}, Y_{1}, Z_{1}, 1)^{\top}$ and $\mathbf{X_{2}} = (X_{2}, Y_{2}, Z_{2}, 1)^{\top}$:

\begin{equation}
    \mathbf{L}_w =
    \begin{bmatrix}
        \mathbf{m}_w \\
        \mathbf{d}_w
    \end{bmatrix}
    =
    \begin{bmatrix}
        Y_{2} Z_{1} - Y_{1} Z_{2} \\
        Z_{2} X_{1} - Z_{1} X_{2} \\
        X_{2} Y_{1} - X_{1} Y_{2} \\
        X_{1} - X_{2} \\
        Y_{1} - Y_{2} \\
        Y_{1} - Y_{2}
    \end{bmatrix}
\end{equation}

The orthonormal representation $(\mathbf{U}, \mathbf{W}) \in SO(3) \times SO(2)$ of a 3D line can be computed using the QR decomposition given the Pl\"ucker coordinates \cite{fu2020plvins}:

\begin{equation}
\left[\mathbf{m}_{w}\mid \mathbf{d}_{w}\right]=\mathbf{U}\begin{bmatrix}\omega_{1} & 0 \\0 & \omega_{2} \\0&0 \end{bmatrix}, with: \mathbf{W}=\begin{bmatrix}\omega_{1} & \omega_{2} \\-\omega_{2} & \omega_{1} \end{bmatrix}
\label{e:qrcomposition}
\end{equation}

\noindent where $\mathbf{U}$ and $\mathbf{W}$ denote a three and a two-dimensional rotation matrix, respectively. Let $\mathbf{R}(\bm{\theta})=\mathbf{U}$ and $\mathbf{R}(\theta)=\mathbf{W}$ be the corresponding rotation transformations, we have:

\begin{equation}
\begin{aligned}
\mathbf{R}(\bm{\theta}) & =\left[\mathbf{u}_{1},\mathbf{u}_{2},\mathbf{u}_{3}\right]=\left[\frac{\mathbf{m}_{w}}{||\mathbf{m}_{w}||},\frac{\mathbf{d}_{w}}{||\mathbf{d}_{w}||},\frac{\mathbf{m}_{w}\times\mathbf{d}_{w}}{||\mathbf{m}_{w}\times \mathbf{d}_{w}||} \right] \\
\mathbf{R}(\theta) & =\begin{bmatrix}\omega_{1} & \omega_{2} \\-\omega_{2} & \omega_{1} \end{bmatrix}=\begin{bmatrix}\cos(\theta) & -\sin(\theta) \\\sin(\theta) & \cos(\theta) \end{bmatrix} \\ 
& =\frac{1}{\sqrt{(||\mathbf{m}_{w}||^{2}+||\mathbf{d}_{w}||^{2})}}\begin{bmatrix}||\mathbf{m}_{w}|| & -||\mathbf{d}_{w}||\\||\mathbf{d}_{w}|| & ||\mathbf{m}_{w}|| \end{bmatrix}
\end{aligned}
\label{e:rotation1}
\end{equation}

\noindent where $\bm{\theta}$ and $\theta$ denote a 3-vector and a scalar, respectively. 

Within the iterative optimization in the BA, $\mathbf{U}$ and $\mathbf{W}$ can be updated as $\mathbf{U} \leftarrow \mathbf{U}\mathbf{R}(\bm{\theta})$ (notice that $\bm{\theta} \in \mathbb{R}^3$) and $\mathbf{W} \leftarrow \mathbf{W}\mathbf{R}(\theta)$ (notice that $\theta \in \mathbb{R}$). Therefore the orthonormal representation has the minimal 4 parameters as $\left[ \bm{\theta}^{\top}, \theta \right] \in \mathbb{R}^{4}$.

Given an orthonormal representation $(\mathbf{U}, \mathbf{W})$, we can recover its Pl\"{u}cker coordinates by:

\begin{equation}
\mathbf{L}_{w}=\left[\omega_{1}\mathbf{u}_{1}^{\top}, \omega_{2}\mathbf{u}_{2}^{\top}\right],
\end{equation}

\noindent where $\omega_{1}$, $\omega_{2}$, $\mathbf{u}_{1}$, and $\mathbf{u}_{2}$ can be extracted from Eq. (\ref{e:rotation1}), as $\mathbf{u}_{i}$ the $i$-th column of $\mathbf{U}$.


\subsection{The Analytical Jacobians of 3D Line}
\label{app: jacobian_line}

The complete Jacobians of reprojection error of the line \cite{lee2019elaborate} regarding to the \textbf{orthonormal representations} (see last Sec. \ref{app: Conversion}) and \textbf{camera poses} $\mathfrak{se}(3)$ are:

\begin{equation}
    \mathbf{J}_{\boldsymbol{\theta}}=\frac{\partial \mathbf{e}_{l}}{\partial \boldsymbol{\delta}_{\boldsymbol{\theta}}}=\frac{\partial \mathbf{e}_{l}}{\partial \mathbf{l}} \frac{\partial \mathbf{l}}{\partial \mathbf{L}_{c}} \frac{\partial \mathbf{L}_{c}}{\partial \mathbf{L}_{w}} \frac{\partial \mathbf{L}_{w}}{\partial \boldsymbol{\delta}_{\boldsymbol{\theta}}}
\end{equation}

\begin{equation}
    \mathbf{J}_{\xi}=\frac{\partial \mathbf{e}_{l}}{\partial \boldsymbol{\delta}_{\xi}}=\frac{\partial \mathbf{e}_{l}}{\partial \mathbf{l}} \frac{\partial \mathbf{l}}{\partial \mathbf{L}_{c}} \frac{\partial \mathbf{L}_{c}}{\partial \boldsymbol{\delta}_{\xi}}
\end{equation}

\noindent where the partial derivative of the line reprojection error w.r.t the reprojected line $\mathbf{l}$ is:

\begin{equation}
    \frac{\partial \mathbf{e}_{l}}{\partial \mathbf{l}}=\frac{1}{\sqrt{l_{1}^{2}+l_{2}^{2}}}\left[\begin{array}{ccc}
x_{s}-\frac{l_{1} \mathbf{x}_{s} \mathbf{l}}{\sqrt{l_{1}^{2}+l_{2}^{2}}} & y_{s}-\frac{l_{2} \mathbf{x}_{s} \mathbf{l}}{\sqrt{l_{1}^{2}+l_{2}^{2}}} & 1 \\
x_{e}-\frac{l_{1} \mathbf{x}_{e} \mathbf{l}}{\sqrt{l_{1}^{2}+l_{2}^{2}}} & y_{e}-\frac{l_{2} \mathbf{x}_{e} \mathbf{l}}{\sqrt{l_{1}^{2}+l_{2}^{2}}} & 1
\end{array}\right]_{2 \times 3}
\end{equation}

\noindent The partial derivatives of the reprojected line $\mathbf{l}$ w.r.t $\mathbf{L}_{c}$, and $\mathbf{L}_{c}$ w.r.t $\mathbf{L}_{w}$:

\begin{equation}
    \begin{aligned}
    \frac{\partial \mathbf{l}}{\partial \mathbf{L}_{c}} &=\frac{\partial \mathbf{K}_{L} \mathbf{m}_{c}}{\partial \mathbf{L}_{c}}=\left[\begin{array}{ll}
    \mathbf{K}_{L} & \mathbf{0}_{3 \times 3}
    \end{array}\right]_{3 \times 6} \\
    \frac{\partial \mathbf{L}_{c}}{\partial \mathbf{L}_{w}} &=\frac{\partial \mathbf{T}_{c w} \mathbf{L}_{w}}{\partial \mathbf{L}_{w}}=\mathbf{T}_{c w} = 
    \begin{bmatrix}
    \mathbf{R}_{cw} & \left[\mathbf{t}_{cw}\right]_{\times}\mathbf{R}_{cw} \\
    \mathbf{0}_{3 \times 3} & \mathbf{R}_{cw} 
    \end{bmatrix}_{6 \times 6}
    \end{aligned}
\end{equation}

\noindent The derivative of $\mathbf{L}_{w}$ w.r.t the orthonormal representation is:

\begin{equation}
    \frac{\partial \mathbf{L}_{w}}{\partial \boldsymbol{\delta}_{\boldsymbol{\theta}}}=\left[\begin{array}{cccc}
    \mathbf{0}_{3 \times 1} & -\omega_{1} \mathbf{u}_{3} & \omega_{1} \mathbf{u}_{2} & -\omega_{2} \mathbf{u}_{1} \\
    \omega_{2} \mathbf{u}_{3} & \mathbf{0}_{3 \times 1} & -\omega_{2} \mathbf{u}_{1} & \omega_{1} \mathbf{u}_{2}
    \end{array}\right]_{6 \times 4}
\end{equation}

\noindent The derivative of $\mathbf{L}_{c}$ w.r.t camera pose is:

\begin{equation}
    \frac{\partial \mathbf{L}_{c}}{\partial \boldsymbol{\delta}_{\xi}}=\left[\begin{array}{cc}
    -[\mathbf{R} \mathbf{m}]_{\times}-\left[[\mathbf{t}]_{\times} \mathbf{Rd}\right]_{\times} & -[\mathbf{R} \mathbf{d}]_{\times} \\
    -[\mathbf{Rd}]_{\times} & \mathbf{0}_{3 \times 3}
    \end{array}\right]_{6 \times 6}
\end{equation}


\subsection{Endpoints Trimming}
\label{app: endpoints_trimming}
Given an observed 2D line segment with endpoints $\{\mathbf{x}_{s}, \mathbf{x}_{e}\}$, and correspondingly a reprojected line $\mathbf{l}$, a perpendicular intersection plane is constructed via finding the closet point $\mathbf{x}_{\perp}$ of $\mathbf{x}_{s}$ (or $\mathbf{x}_{e}$) to the line $\mathbf{l}$ \cite{lee2019elaborate, zhang2015building}:

\begin{equation}
x_{\perp s} = - \left(y_{s} - \frac{l_{2}}{l_{1}}x_{s} + \frac{l_{3}}{l_{2}}\right)\frac{l_{1}l_{2}}{l_{1}^{2} + l_{2}^{2}} 
\end{equation}

\begin{equation}
y_{\perp s} = - \frac{l_{1}}{l_{2}}x_{\perp s} - \frac{l_{3}}{l_{2}}
\end{equation}

Calculating a random point $\mathbf{x_{0s}}$, e.g., with $x_{0s} = 0$:

\begin{equation}
    x_{0s} = 0, y_{0s} = y_{s} - \frac{l_{2}}{l_{1}}x_{s}
\end{equation}

By doing that, we could construct a 3D plane by:

\begin{equation}
    \boldsymbol{\pi}_{s} = \mathbf{P}^{\top}\mathbf{l}_{\perp s}
\end{equation}

\noindent where $\mathbf{l}_{\perp s} = \mathbf{x}_{\perp s} \times \mathbf{x_{0s}}$. Given the Pl\"ucker coordinates of the 3D line, we are able to calculate the 3D starting point via intersecting the 3D plane and 3D line \cite{hartley2003multiple} via $ \mathbf{X}_{s} = (X_{s}/\omega,Y_{s}/\omega,Z_{s}/\omega, 1)^{\top} = \mathbf{L}\boldsymbol{\pi}_s$, where:

\begin{equation}
    \mathbf{L} = 
    \begin{bmatrix}
    [\mathbf{m}]_{\times} & \mathbf{d} \\
    -\mathbf{d}^{\top} & 0
    \end{bmatrix}
\end{equation}

The ending point of the 3D line is estimated by the same procedure described above for the starting point. 


\begin{table}[!t]
\centering
\scalebox{0.9}{
\begin{tabular}{|l||c|c|c|}
\hline
    \textbf{Monocular} & \textbf{Ours} & \textbf{Ours} & OpenVSLAM \cite{sumikura2019openvslam}\\
                       & (full)        &               &\\
\hline
    Config. & point   & point   & point   \\
            & + line  & + line  &         \\
            & + plane &         & \\
\hline
\hline
    MH\_01\_easy   &-	&0.043	&0.045\\
    MH\_02\_easy   &-	&0.035	&0.037\\
    MH\_03\_medium &-	&0.038	&0.039\\
    MH\_04\_difficult &-	&0.155	&0.177\\
    MH\_05\_difficult &-	&0.088	&0.075\\
\hline
    Avg. &-	&\textbf{0.072}	&0.074\\
\hline
\hline
    V1\_01\_easy &0.093	&0.096	&0.095\\
    V1\_02\_medium &0.062	&0.065	&0.064\\
    V1\_03\_difficult &0.070	&0.069	&0.069\\
    V2\_01\_easy &0.058	&0.059	&0.063\\
    V2\_02\_medium &0.057	&0.057	&0.064\\
    V2\_03\_difficult &0.151	&0.106	&0.127\\
\hline
    Avg. &0.082	&\textbf{0.075}	&0.080\\
\hline
\end{tabular}
}
\caption{\textbf{Monocular SLAM} evaluated on dataset \textbf{EuRoC MAV} \cite{Burri25012016}, presented are the \textbf{absolute trajectory errors (ATEs) RMSE [m]} (- stands result not available due to the failure of instance planar segmentation CNN on factory image sequences \textit{MH\_01 - 05}).  Each result from ours was calculated as the average of over 5 executions.}
\label{tab: ATE-euroc-mono}
\end{table}
\begin{table}[!t]
\centering
\scalebox{0.9}{
\begin{tabular}{|l||c|c|c|}
\hline
    \textbf{Stereo} & \textbf{Ours} & \textbf{Ours} & OpenVSLAM \cite{sumikura2019openvslam}\\
                       & (full)        &               &\\
\hline
    Config. & point   & point   & point   \\
            & + line  & + line  &         \\
            & + plane &         & \\
\hline
\hline
    MH\_01\_easy   &-	&0.046	&0.050\\
    MH\_02\_easy   &-	&0.056	&0.058\\
    MH\_03\_medium &-	&0.048	&0.053\\
    MH\_04\_difficult &-	&0.071	&0.072\\
    MH\_05\_difficult &-	&0.071	&0.064\\
\hline
    Avg. &-	&0.059	&0.059\\
\hline
\hline
    V1\_01\_easy &0.089	&0.091	&0.093\\
    V1\_02\_medium &0.066	&0.066	&0.067\\
    V1\_03\_difficult &0.064	&0.065	&0.073\\
    V2\_01\_easy &0.060	&0.061	&0.061\\
    V2\_02\_medium &0.060	&0.061	&0.064\\
    V2\_03\_difficult &0.163	&0.166	&0.157\\
\hline
    Avg. &\textbf{0.084}	&0.085	&0.086\\
\hline
\end{tabular}
}
\caption{\textbf{Stereo SLAM} evaluated on dataset \textbf{EuRoC MAV} \cite{Burri25012016}, presented are the \textbf{absolute trajectory errors (ATEs) RMSE [m]}.}
\label{tab: ATE-euroc-stereo}
\end{table}
\begin{table}[!t]
\centering
\scalebox{0.7}{
 \begin{tabular}{|l|c|c|c|}
    \hline 
        \textbf{Thread} & \textbf{Ours} & ORB-SLAM2 \cite{mur2017orb} & OpenVSLAM \cite{sumikura2019openvslam} \\
    \hline 
    \hline
        Tracking & 28.41 & 20.43 & 19.47 \\
        Local Mapping & 269.08 & 110.82 & 105.45 \\
    \hline
    \hline
    \textbf{Functionality/Module} & \multicolumn{3}{c|}{\textbf{Ours}} \\
    \hline
    \hline
    System (Mono) Initialization & \multicolumn{3}{c|}{5.37} \\
    \hline
    2D ORB Feature Extraction & \multicolumn{3}{c|}{11.39} \\
    2D Line Feature Extraction & \multicolumn{3}{c|}{13.70} \\
    Track Local Map (point + line) & \multicolumn{3}{c|}{12.69} \\
    Motion-only BA (point + line) & \multicolumn{3}{c|}{5.45} \\
    \hline
    Instance Planar Segmentation & \multicolumn{3}{c|}{69.45} \\
    Point-Plane Map Initialization & \multicolumn{3}{c|}{22.14} \\
    Non-Planar Map Point Culling & \multicolumn{3}{c|}{0.56} \\
    New Plane Detection & \multicolumn{3}{c|}{2.07} \\
    Map Plane Merging/Expanding & \multicolumn{3}{c|}{0.92} \\
    Map Plane Re-estimation & \multicolumn{3}{c|}{1.11} \\
    Map Point-Plane Refinement & \multicolumn{3}{c|}{0.08} \\
    \hline
    Map Point Two-keyframe Triangulation & \multicolumn{3}{c|}{0.01} \\
    Map Line Two-keyframe Triangulation & \multicolumn{3}{c|}{0.21} \\
    Map Point Culling & \multicolumn{3}{c|}{0.87} \\
    Map Line Culling & \multicolumn{3}{c|}{0.10} \\
    Local BA (point + line + endpoints trimming) & \multicolumn{3}{c|}{233.37} \\
    \hline
    \end{tabular}
}
\caption{\textbf{Runtime analysis [ms] (mean value evaluated on dataset TUM RGB-D \cite{sturm2012benchmark}: fr3\_st\_tex\_far)} of our full SLAM system compared to original ORB-SLAM2 and OpenVSLAM, \textbf{under monocular setting}, using a desktop PC with an Intel Xeon(R) E-2146G 12 cores CPU @ 3.50GHz, 32GB RAM. The PlaneRecNet \cite{xie2021planerecnet} is evaluated on a standard GPU of NVIDIA GTX 1650.}
\label{tab: run-time}
\end{table}


\begin{figure}[!t]
    \centering
    \subfigure[\textbf{Point-line} map reconstructed from our \textbf{monocular SLAM}, of data sequence \textit{living\_room\_traj0}.]{\includegraphics[width=0.85\linewidth]{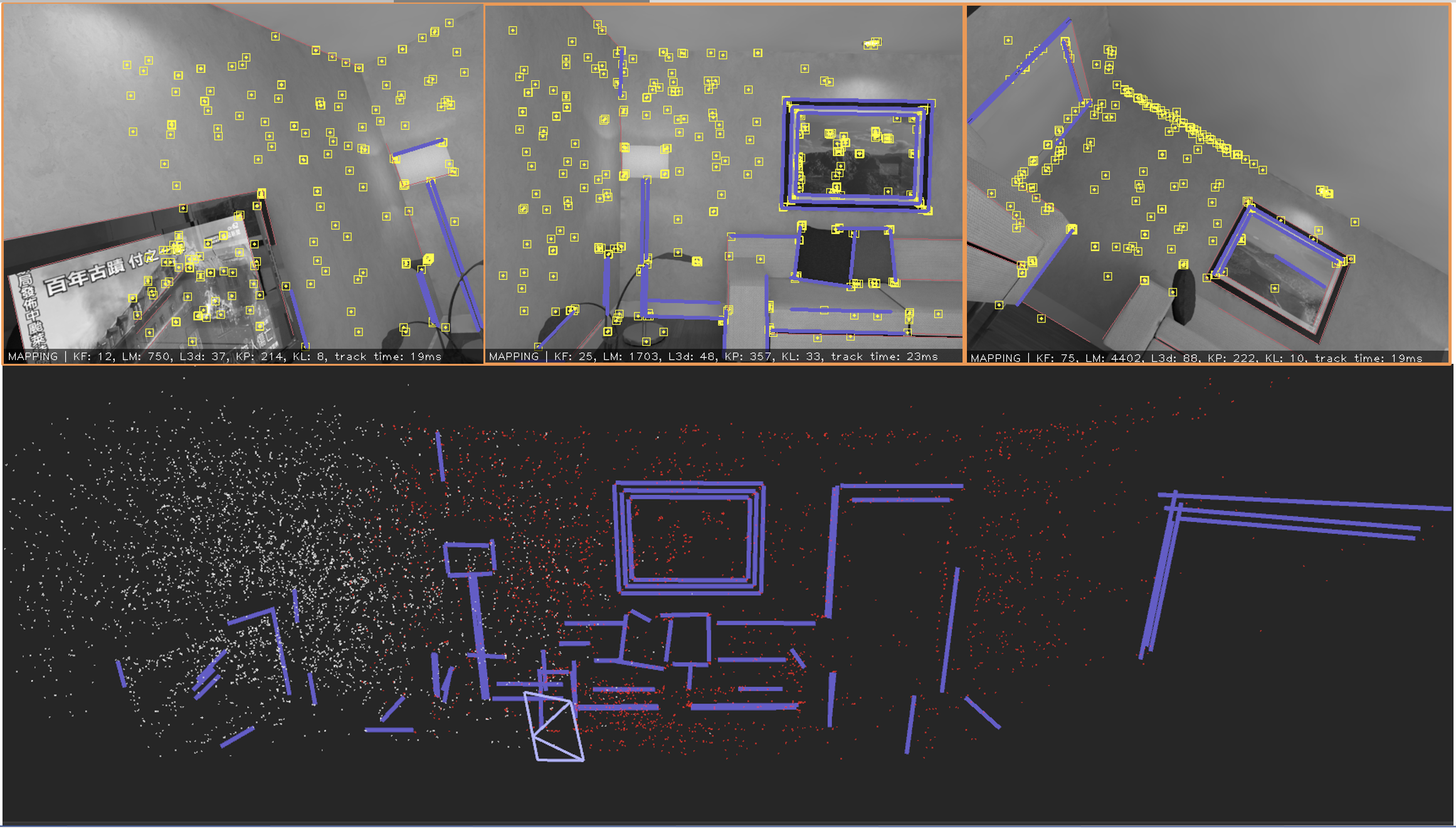}}
    
    \subfigure[\textbf{Line-plane} map (middle) and \textbf{point-line} map  reconstructed from our \textbf{monocular SLAM}, of data sequence \textit{living\_room\_traj2}.]{\includegraphics[width=0.85\linewidth]{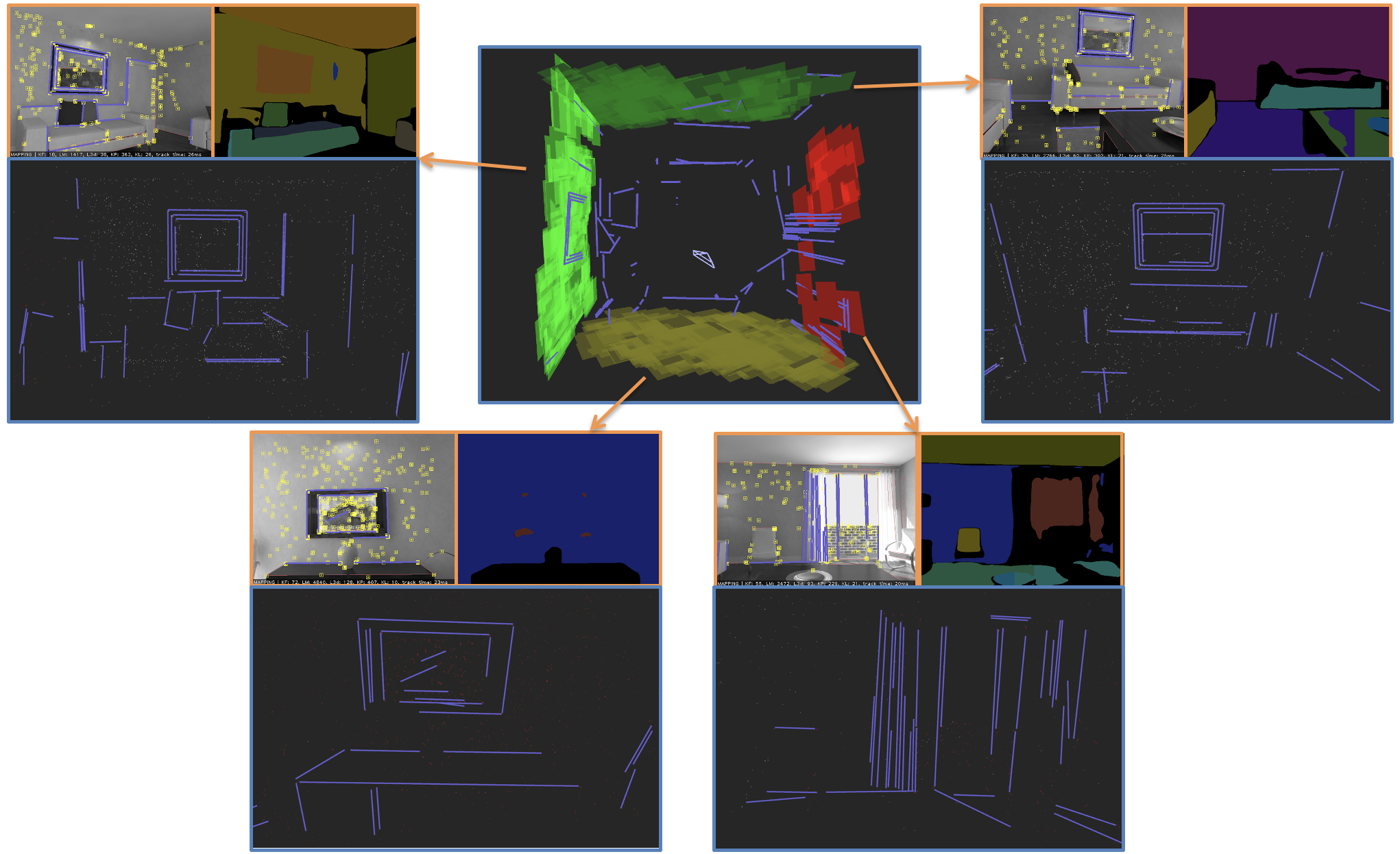}}

    \caption{\textbf{A qualitative illustration} of the map reconstructed from our \textbf{monocular SLAM}, on ICL-NUIM dataset \cite{handa2014benchmark}.}
    \label{fig: qualitative-mono-more}
\end{figure}


\subsection{More Quantitative and Qualitative Results}
\label{app: more_evaluation}

\textbf{Dataset EuRoC MAV.} As presented in Table \ref{tab: ATE-euroc-mono} and Table \ref{tab: ATE-euroc-stereo}, we tested our monocular and stereo SLAM on dataset EuRoC MAV \cite{Burri25012016} which consists of 11 stereo sequences recorded with a MAV flying across three different environments: two indoor rooms and one industrial scenario, containing sequences that present different challenges depending on the speed of the drone, illumination, texture, etc. Note that results of exploiting planar reconstruction are not available on factory image sequences \textit{MH\_01 - 05}, due to the failure of instance planar segmentation CNN. 

The exploitation of line segments barely increases the accuracy on dataset EuRoC MAV, similar results were also mentioned in stereo PL-SLAM \cite{gomez2019pl}. Thus, stereo PL-SLAM only reports the relative pose errors of the keyframes, which are not compared in Table \ref{tab: ATE-euroc-stereo} using ATEs. 
Still, our SLAM system shows slightly superior performance for most of the sequences in comparison to the point-only approach, especially the reconstructed map is more intuitive.

\begin{figure}[!t]
    \centering
    \subfigure[Point-line map reconstructed from our \textbf{monocular SLAM}.]{\includegraphics[width=0.9\linewidth]{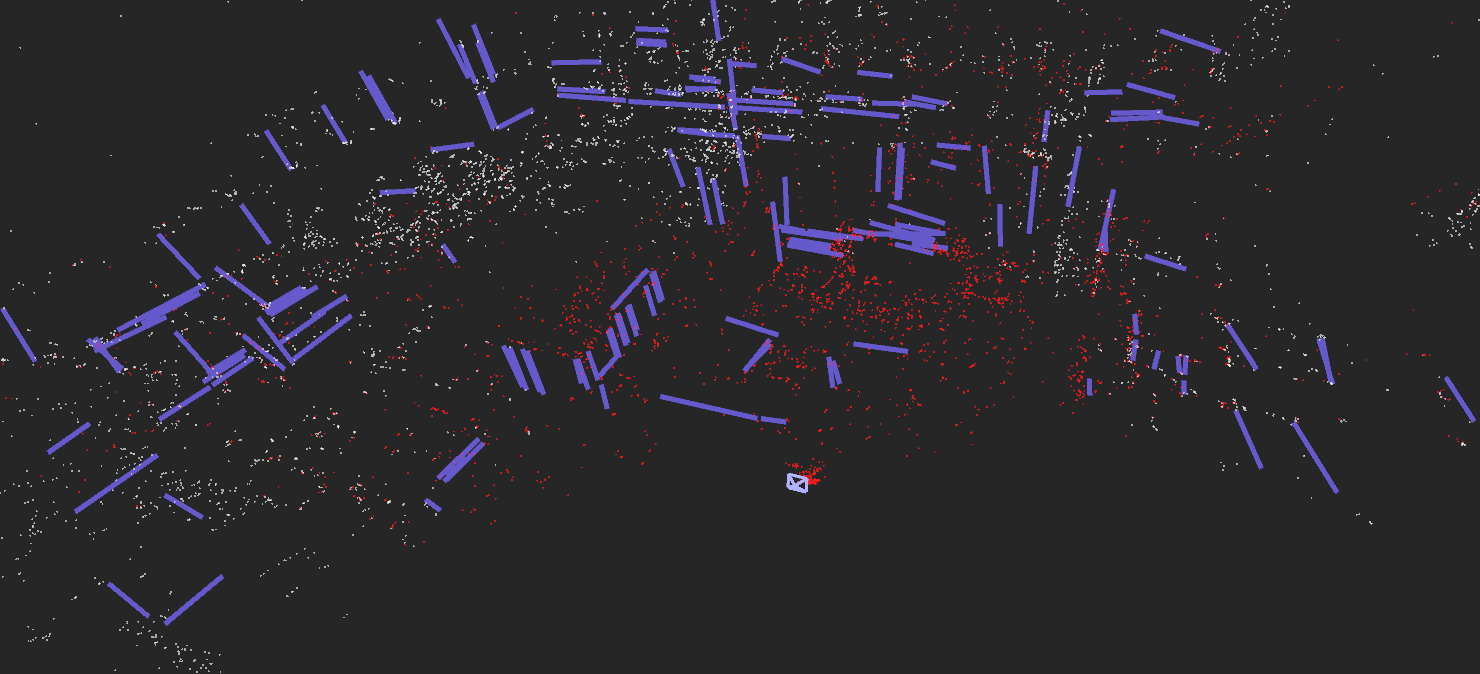}}
    
    \subfigure[Point-line map reconstructed from our \textbf{stereo SLAM}.]{\includegraphics[width=0.9\linewidth]{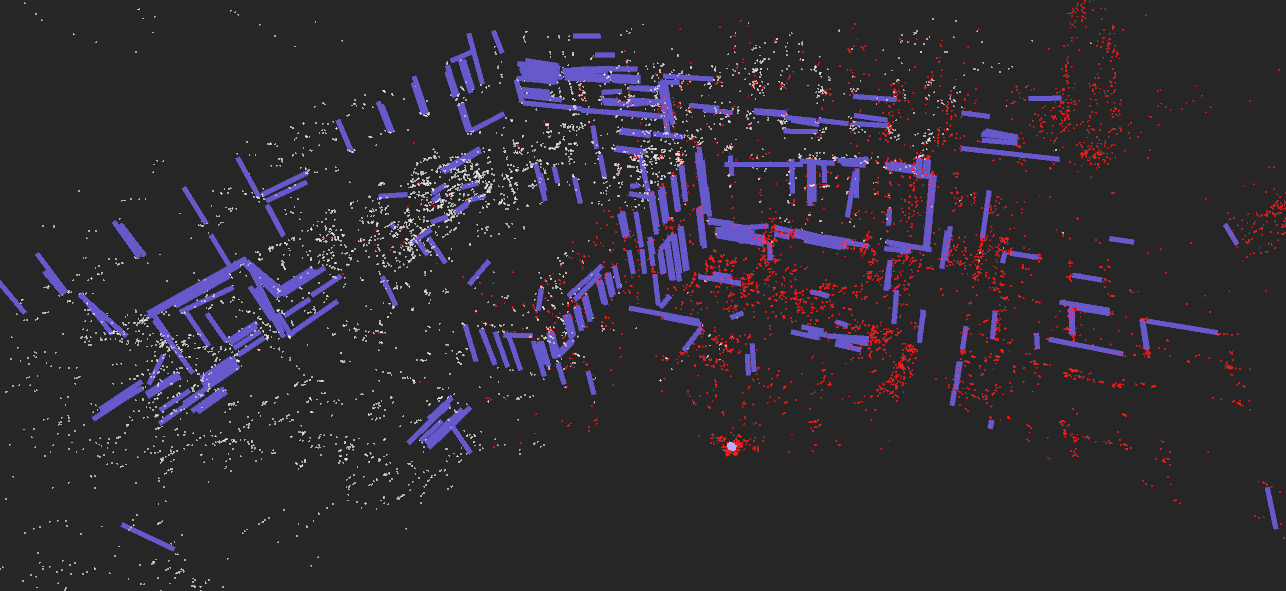}}
    
    \subfigure[Point-line map reconstructed from \textbf{PL-VINS}.]{\includegraphics[width=0.9\linewidth]{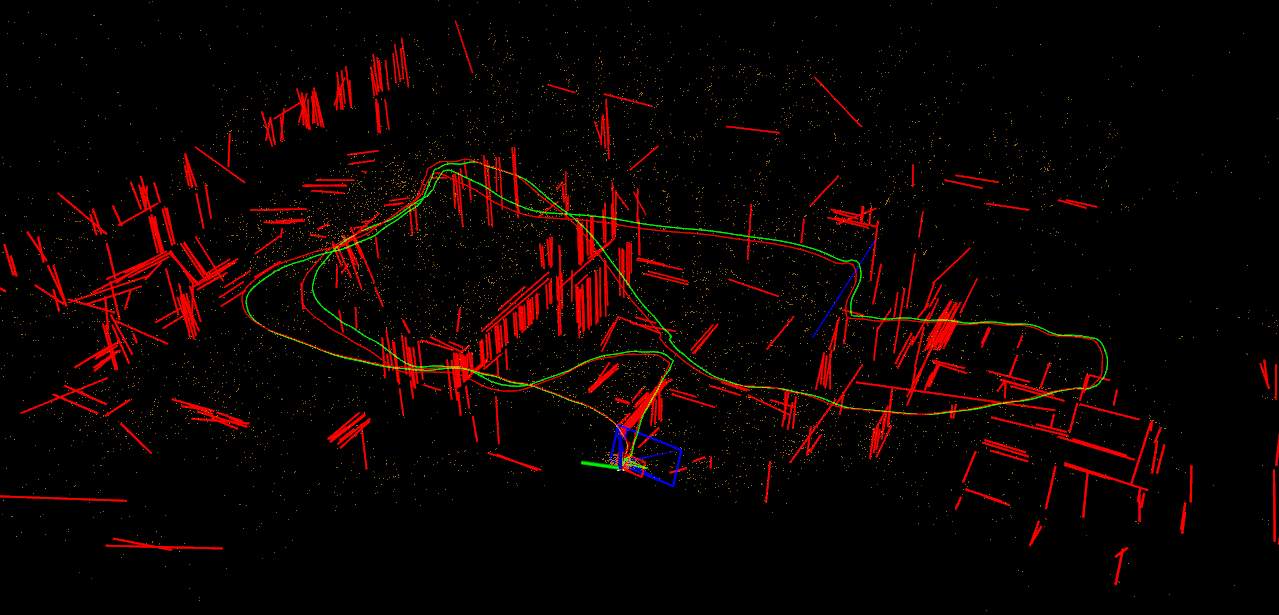}}
    
    \caption{\textbf{A qualitative comparison between our stereo SLAM and PL-VINS} \cite{fu2020plvins}, on \textit{MH\_04\_difficult} of EuRoC dataset \cite{Burri25012016}.}
    \label{fig: qualitative-vins}
\end{figure}

\begin{figure}[!t]
    \centering
    \subfigure[Point cloud map reconstructed from our \textbf{stereo SLAM}.]{\includegraphics[width=0.9\linewidth]{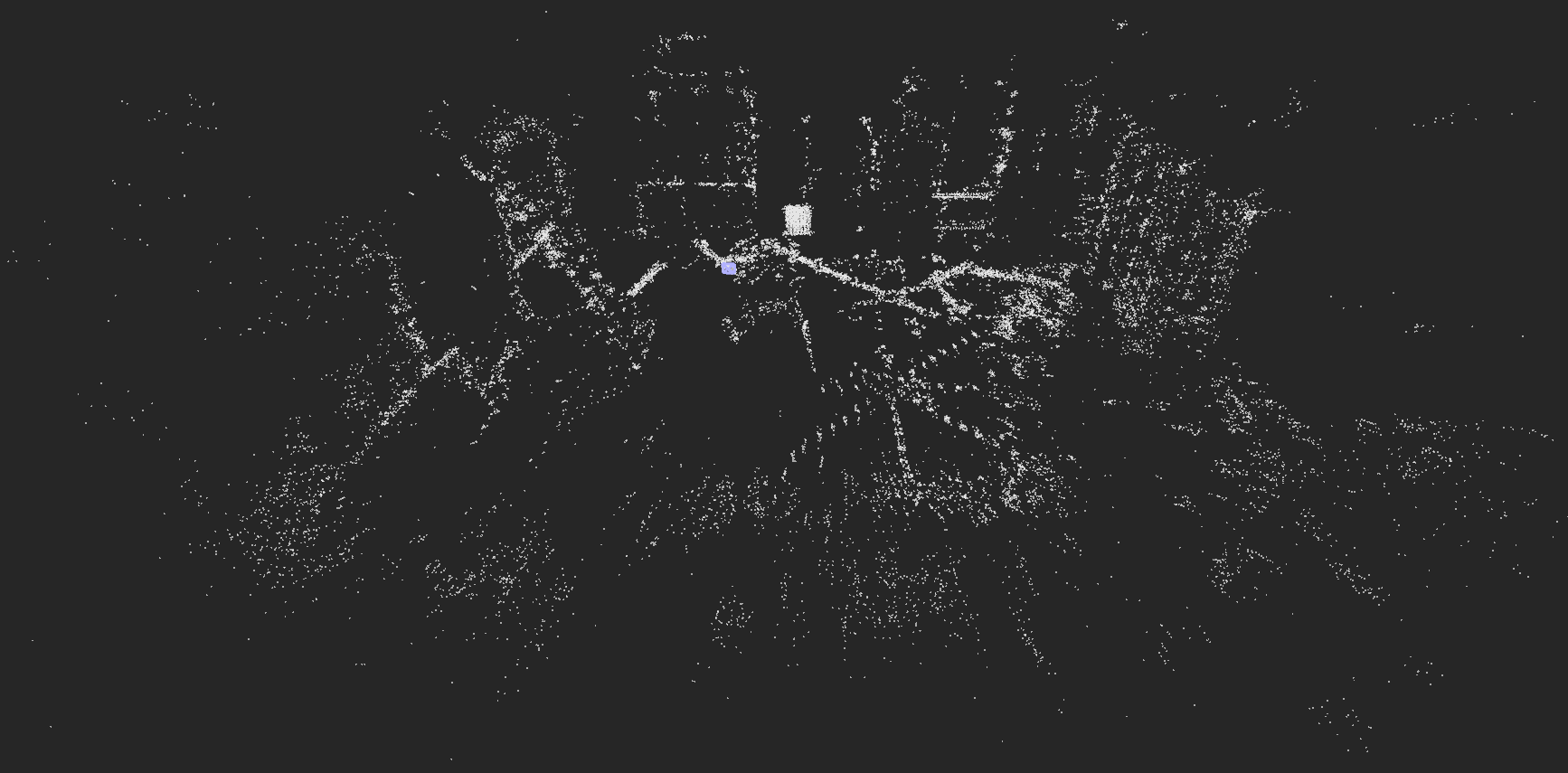}}
        
    \subfigure[Line-plane map reconstructed from our \textbf{stereo SLAM}.]{\includegraphics[width=0.9\linewidth]{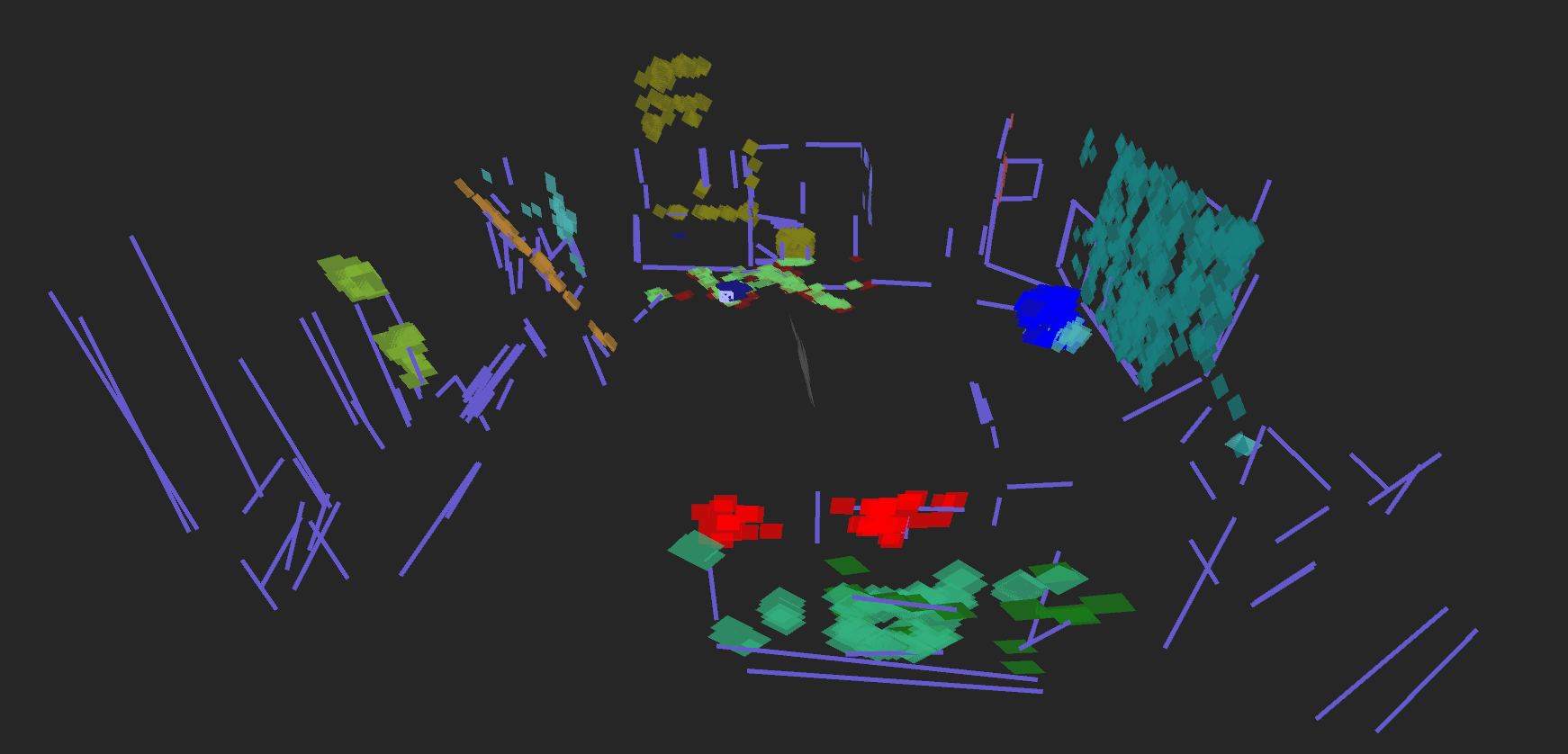}}
    
    \subfigure[Full visualization of map reconstructed from our \textbf{stereo SLAM}.]{\includegraphics[width=0.9\linewidth]{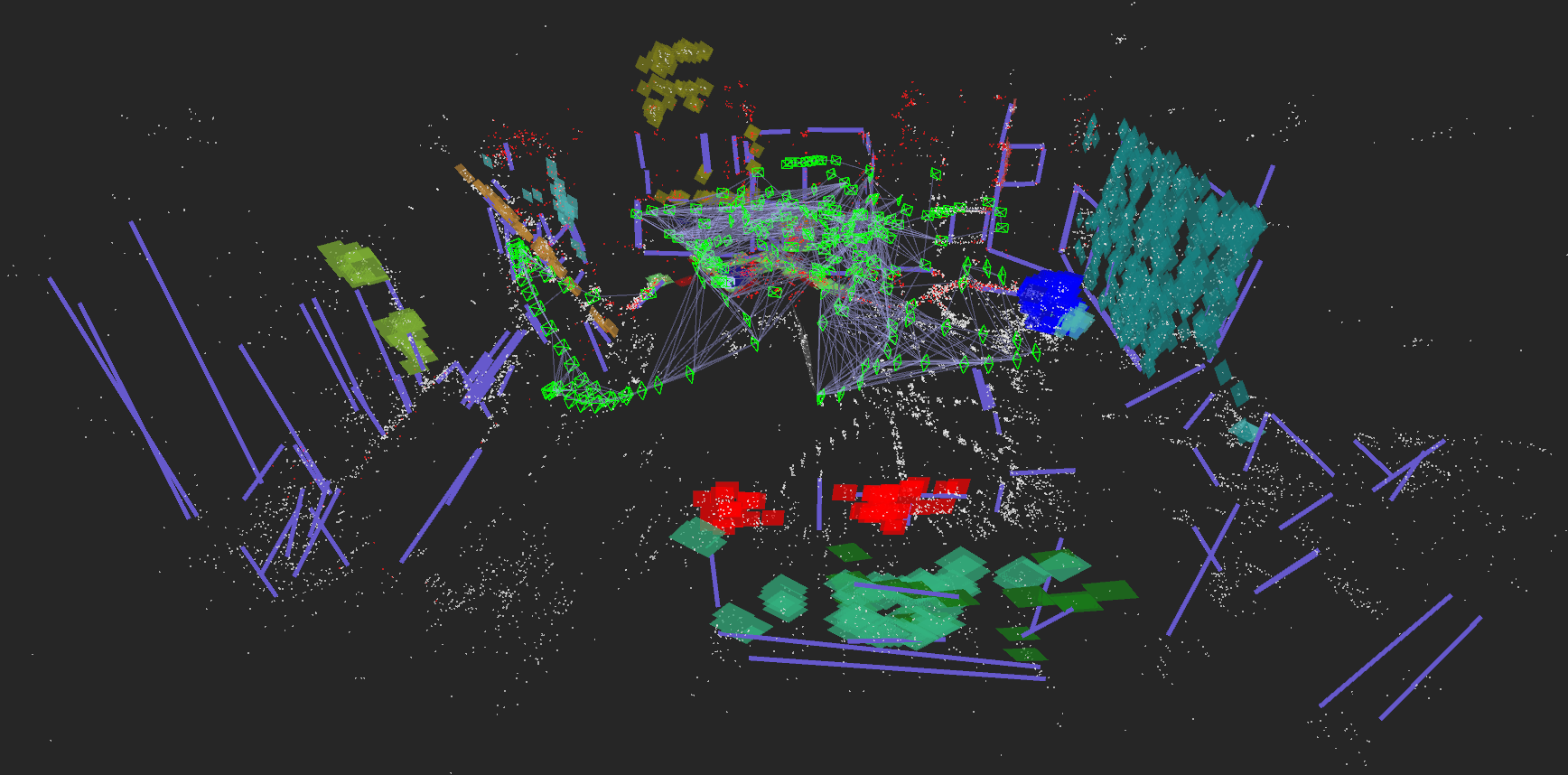}}
    \caption{\textbf{A qualitative illustration} of our \textbf{stereo SLAM}, on \textit{V1\_03\_difficult} of EuRoC dataset \cite{Burri25012016}.}
    \label{fig: qualitative-euroc-more}
\end{figure}

\textbf{More qualitative illustration.} More examples from our monocular SLAM on the ICL-NUIM dataset see Fig. \ref{fig: qualitative-mono-more}. Another example of our RGB-D SLAM is given in Fig. \ref{fig: qualitative-rgbd-more}. As mentioned in the main paper, we also provide a comparison between ours and PL-VINS \cite{fu2020plvins} in Fig. \ref{fig: qualitative-vins}, qualitatively showing that both our monocular and stereo SLAM provide highly accurate point and line maps. Another example with planar reconstruction is given in Fig. \ref{fig: qualitative-euroc-more}. 

\textbf{Computation Complexity.} A detailed run-time analysis (in ms) is given in Table \ref{tab: run-time}, of which our full monocular SLAM system utilizes point and line features in the tracking thread, and reconstructs 3D points, 3D planes, and 3D lines in the local mapping thread. 

One can observe that the tracking thread of our PLP-SLAM needs more time on average (about 28ms) compared to the original ORB-SLAM2 (about 20ms), but it can still achieve a tracking performance of 30 frames per second. Notice that several functionalities are implemented as multi-thread processing, for example, the ORB and LSD features are extracted in parallel threads. The mapping thread costs more time due to the extended local BA with lines (about 230ms), while new 3D plane fitting (about 2ms), merging (about 0.9ms), and refinement (about 1ms) are in general very fast. During the experiments, we do not observe an obvious delay in the mapping visualization. Hence, most of the added computation burden is from the utilization of the line feature. For interested readers, please note that the results given in Table \ref{tab: run-time} may vary slightly if the hardware configuration (e.g. CPU) changes, and may vary slightly on different data image sequences (e.g. in the case of various sizes of the map).

\subsection{Remark}

\begin{figure}[!t]
    \centering
    \includegraphics[width=0.85\linewidth]{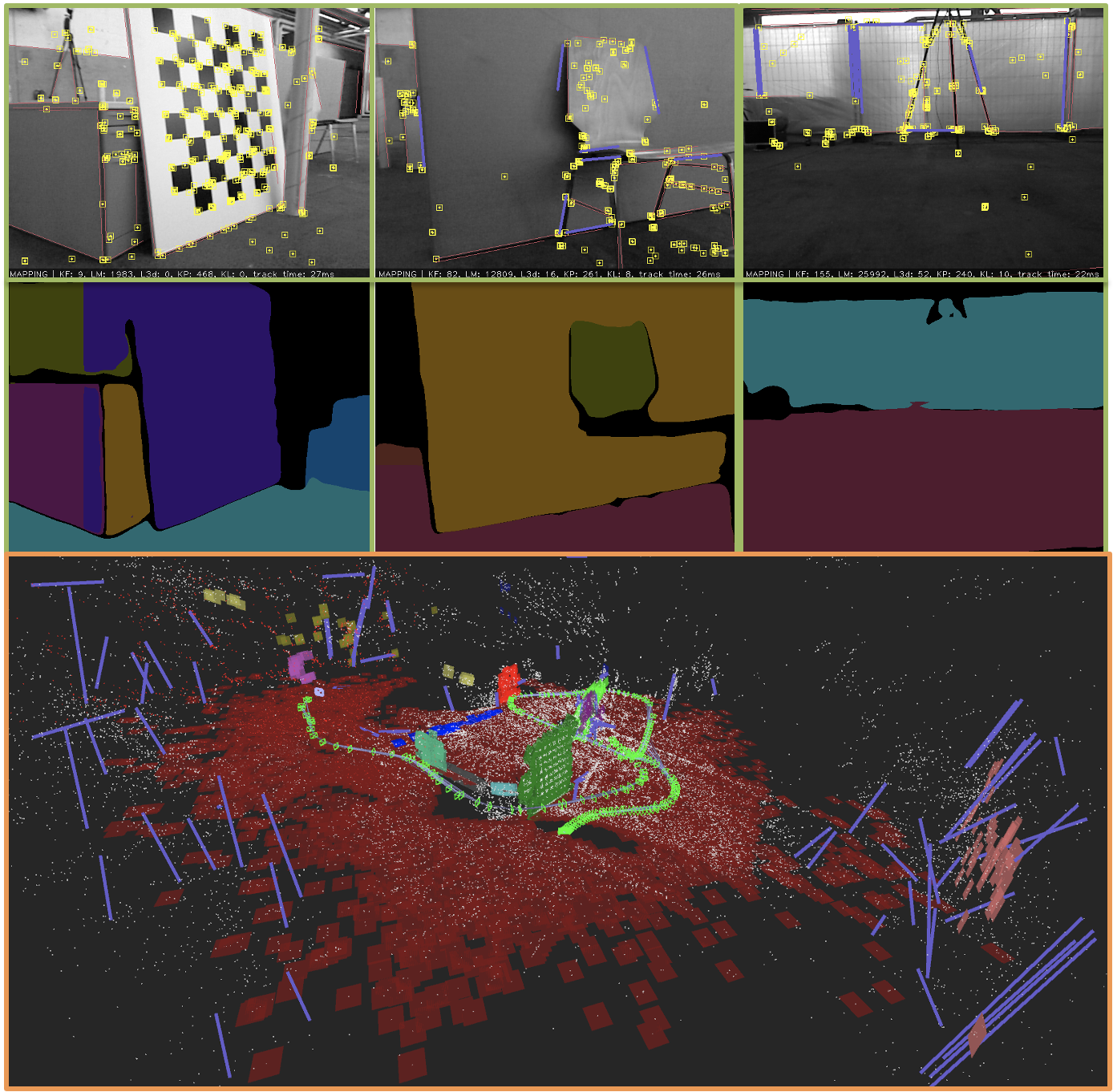}

    \caption{\textbf{A qualitative illustration} of the \textbf{point-line-plane} map reconstructed from our \textbf{RGB-D SLAM}, on TUM RGB-D dataset \cite{sturm2012benchmark}, of data sequence \textit{fr2\_pioneer\_slam}.}
    \label{fig: qualitative-rgbd-more}
\end{figure}

As concluded in the main paper, the plane detection and reconstruction algorithms are mainly designed for monocular settings. Therefore, the algorithms can be applied under RGB-D or stereo camera settings, but are not necessarily the optimal methods, since more information can be used to detect and reconstruct planes when available (e.g. depth image). Nevertheless, we found out the designed algorithms can provide an intuitive map under RGB-D settings in low-texture scenes, as a result of sparse semantic mapping, for example, see Fig. \ref{fig: qualitative-rgbd-more}.















\end{document}